\documentclass[10pt,journal,compsoc]{IEEEtran}

\usepackage{amsmath}
\usepackage{amsfonts}
\usepackage{amssymb}
\usepackage{bbm}
\usepackage{graphicx}
\graphicspath{{graphics/}}
\usepackage{color}
\usepackage{url}
\usepackage{footnote}
\makesavenoteenv{tabular}
\usepackage{algorithm}
\usepackage{algorithmic}
\usepackage{subfigure}
\usepackage{tabularx}
\usepackage{tabularx,booktabs}
\usepackage{multirow}
\usepackage{url}
\usepackage[switch]{lineno}
\usepackage{xcolor}

\ifCLASSOPTIONcompsoc
  \usepackage[nocompress]{cite}
\else
  \usepackage{cite}
\fi

\ifCLASSINFOpdf
\else
\fi
\begin{document}

\title{Bayesian Temporal Factorization for Multidimensional Time Series Prediction}
%
%
%

\author{Xinyu~Chen 
        and~Lijun~Sun,~\IEEEmembership{Member,~IEEE}
\IEEEcompsocitemizethanks{
\IEEEcompsocthanksitem Xinyu Chen is with the Department  of  Civil, Geological and  Mining  Engineering, Polytechnique Montreal, Montreal, QC, H3T 1J4, Canada. E-mail: chenxy346@gmail.com
\IEEEcompsocthanksitem  Lijun Sun is with the Department of Civil Engineering, McGill University, Montreal, QC, H3A 0C3, Canada. E-mail: lijun.sun@mcgill.ca
\IEEEcompsocthanksitem Lijun Sun is also with the Interuniversity Research Centre on Enterprise Networks, Logistics and Transportation (CIRRELT), Montreal, QC, H3T 1J4, Canada.}
\thanks{Manuscript received XXX; revised YYY. (Corresponding author: Lijun Sun)}}

\IEEEtitleabstractindextext{%
\begin{abstract}

Large-scale and multidimensional spatiotemporal data sets are becoming ubiquitous in many real-world applications such as monitoring urban traffic and air quality. Making predictions on these time series has become a critical challenge due to not only the large-scale and high-dimensional nature but also the considerable amount of missing data. In this paper, we propose a Bayesian temporal factorization (BTF) framework for modeling multidimensional time series---in particular spatiotemporal data---in the presence of missing values. By integrating low-rank matrix/tensor factorization and vector autoregressive (VAR) process into a single probabilistic graphical model, this framework can characterize both global and local consistencies in large-scale time series data. The graphical model allows us to effectively perform probabilistic predictions and produce uncertainty estimates without imputing those missing values. We develop efficient Gibbs sampling algorithms for model inference and model updating for real-time prediction and test the proposed BTF framework on several real-world spatiotemporal data sets for both missing data imputation and multi-step rolling prediction tasks. The numerical experiments demonstrate the superiority of the proposed BTF approaches over existing state-of-the-art methods.

\end{abstract}
}

\maketitle
\IEEEdisplaynontitleabstractindextext
\IEEEpeerreviewmaketitle


\section{Introduction}

With recent advances in sensing technologies, large-scale and multidimensional time series data---in particular spatiotemporal data---are collected on a continuous basis from various types of sensors and applications. Making predictions on these time series, such as forecasting urban traffic states and regional air quality, serves as a foundation to many real-world applications and benefits many scientific fields \cite{faloutsos2018forecasting,shi2018machine}. For example, predicting the demand and states (e.g., speed, flow) of urban traffic is essential to a wide range of applications in intelligent transportation systems (ITS), such as trip planning, travel time estimation, route planning, traffic signal control, to name but a few \cite{li2018briefoverview}. However, given the complex spatiotemporal dependencies in these data sets, making efficient and reliable predictions for real-time applications has been a long-standing and fundamental research challenge.

Despite the vast body of literature on time series analysis from many scientific areas, three emerging issues in modern sensing technologies have posed challenges to classical modeling frameworks. First, modern time series data are often large-scale, collected from a large number of subjects/locations/sensors simultaneously. For example, the highway traffic Performance Measurement System (PeMS) in California consists of more than 35,000 detectors, and it has been recording flow and speed information every 30 seconds since 1999 \cite{chen2001freeway}. However, most classical time series models, such as vector autoregrssive models (VAR) \cite{hyndman2018forecasting}, are not scalable to handle these large data sets. Second, modern time series generated by advanced sensing technologies are usually high-dimensional with different attributes. The multidimensional property in these time series data sets makes it very difficult to characterize the higher-order correlations/dependencies together with the temporal dynamics across different dimensions \cite{jing2018highorder}. In addition to sensing data, multidimensional time series is also ubiquitous in social science domains such as international relations \cite{pmlr-v48-schein16}, dynamic import-export networks and social networks \cite{chen2019factor}. It is also particularly important in modeling traffic/transportation systems with both origin and destination attributes. For example, mobility demand/flow for different types of travelers using different modes can be modeled as a third (origin zone$\times$destination zone$\times$time) tensor time series and all dimensions have strong interactions with each other \cite{sun2016understanding}. Third, most existing time series models require complete time series data as input, while in real-world time series data sets the missing data problem is almost inevitable due to various factors such as hardware/software failure, human error, and network communication problems. Taken together, it has become a critical challenge to perform reliable forecasting on large-scale time series data in the presence of missing data \cite{anava2015online}. A simple and natural solution is to adopt a two-stage approach: first applying imputation algorithms to fill in those missing entries, and then performing predictions based on the complete time series. This simple two-stage approach has been used in a wide range of real-world applications \cite{che2018recurrent}; however, by applying imputation first, the prediction task actually suffers from accumulated errors resulting from the imputation algorithm.

To address these issues in modeling multivariate and multidimensional time series data, several notable approaches have been proposed recently based on matrix/tensor factorization (see \cite{faloutsos2018forecasting} for a brief review). As a common technique for collaborative filtering, matrix/tensor factorization presents a natural solution to address the scalability, efficiency, and missing data issues. Essentially, these models assume that the multivariate and multidimensional time series can be characterized by a low-rank structure with shared latent factors (i.e., global consistency). In order to create meaningful temporal patterns, different smoothing techniques and regularization schemes have been introduced (e.g., linear dynamical systems \cite{xiong2010temporalcf} and Gaussian processes \cite{luttinen2009variational,nguyen2014collaborative}) to encode local consistency. In a recent work, Yu \emph{et al.} \cite{yu2016temporal} proposed a Temporal Regularized Matrix Factorization (TRMF) framework to model multivariate time series with missing data by introducing a novel AR regularizer on the temporal factor matrix. This work is further extended in Takeuchi \emph{et al.} \cite{takeuchi2017autoregressive} to model spatiotemporal tensor data by introducing a spatial autoregressive regularizer, which provides additional predictive ability (i.e., kriging) on the spatial dimension for unknown locations/sensors.

Overall, these factorization approaches have shown superior performance in modeling real-world large-scale time series data in the presence of missing values; however, there are still several main drawbacks hindering the application of these models. On the one hand, these models in general require careful tuning of the regularization parameters to ensure model accuracy and to avoid overfitting. The tuning procedure is computationally expensive and the cost increases exponentially with the number of parameters. Despite its high computational cost, the tuning procedure has to be performed for each specific study/task/data set and there exist no universal solutions. On the other hand, most
most existing models are either not probabilistic \cite{yu2016temporal} (thus they can only provide point estimates of the time series data) or only designed for imputation/interpolation tasks \cite{xiong2010temporalcf,luttinen2009variational}. As a result, the reliability and uncertainty of the predictions/imputations are often overlooked. However, emerging real-world applications, such as route planning and travel time estimation, are extremely sensitive to uncertainties and risks.

In this paper, we propose a new Bayesian Temporal Factorization (BTF) framework which can effectively handle both the missing data problem and the large-scale/high-dimensional properties in modern spatiotemporal data. Our fundamental assumption is that these time series are highly correlated with shared latent factors. Inspired by the recent studies on temporal regularization \cite{yu2016temporal} and Bayesian factorization \cite{xiong2010temporalcf}, this framework applies low-rank matrix/tensor factorization to model multivariate and multidimensional spatiotemporal data and imposes a vector autoregressive (VAR) process to model the temporal factor matrix. The overall contribution of this framework is threefold:
\begin{enumerate}
    \item We integrate VAR and probabilistic matrix/tensor factorization into a single graphical model to efficiently and effectively model large-scale and multidimensional (spatiotemporal) time series. This model can impute missing values and make prediction without introducing potential bias. By introducing a proper VAR process, we can better capture the dependencies among different temporal factors and handle the case of extreme data corruption.
    \item The framework is fully Bayesian and free from tuning regularization parameters, and thus it gives a flexible solution to ensure model accuracy while avoiding overfitting. By using conjugate priors, we can derive efficient Markov chain Monte Carlo (MCMC) sampling algorithm for model inference. The Bayesian framework allows us to make probabilistic predictions with uncertainty estimates.
    \item  Extensive experiments are performed on real-world spatiotemporal data sets to demonstrate its effectiveness against recent state-of-the-art models.
\end{enumerate}

The rest of this paper is organized as follows. In Section~\ref{sec:review}, we briefly review related work on modeling multivariate time series data and matrix/tensor factorization models for large-scale and multidimensional time series data. Section~\ref{sec:problem_description} provides a detailed description of the multivariate and multidimensional time series prediction problem in the presence of missing values. In Section~\ref{sec:btmf_model}, we present the Bayesian Temporal Matrix Factorization (BTMF) model for matrix time series data and develop an efficient MCMC algorithm for model inference. Section~\ref{sec:bttf_model} extends BTMF to Bayesian Temporal Tensor Factorization (BTTF) to model tensor time series data. Section~\ref{sec:experiments} provides the results on extensive numerical experiments based on several real-world spatiotemporal data sets, followed by the conclusion and future work in Section~\ref{sec:conclusion}.

\section{Related Work} \label{sec:review}

{

In this section, we review and summarize some related studies on modeling multiverse time series data $Y\in \mathbb{R}^{N\times T}$, where $N$ is the number of time series and $T$ is number of time points. We are particularly interested in the case where $N$ is large and $Y$ contains missing values, where it becomes infeasible to apply traditional methods such as VAR (with $\sim N^2$ parameters). We first review a popular solution that models the original data on a much smaller latent space with $R$ factors ($R\ll N$) and then imposes smoothness/dynamics on the temporal latent factors. Modeling multivariate time series data in a latent space also provides a natural solution to solve the missing data problem. Afterwards, we review relevant studies on probabilistic and Bayesian matrix/tensor factorization, which is also a key component of our work.

\subsection{Modeling in Latent Space with Temporal Dynamics}

As mentioned, we can consider the observed data $Y$ an incomplete matrix, on which we can perform dimensionality reduction using matrix factorization (MF) techniques ($Y=W^{\top}X$ with $W\in\mathbb{R}^{R \times N}$ and $X\in\mathbb{R}^{R \times T}$). However, the default MF models only capture the global low rank structure, and thus the results are invariant to the permutation of rows and columns. This is clearly not the case for time series data, as we expect strong local patterns and consistency (e.g., observations at $t$ and $t+1$ should be strongly correlated, and $X$ should be smooth over time). We next review some recent and representative studies on introducing temporal smoothness and temporal dynamics into the lower-dimensional latent variable model.

A central challenge in building latent variable models is to design appropriate regularization terms to incorporate temporal dynamics and smoothness, with the goal to both achieve high accuracy and avoid overfitting. On this track, many studies have introduce regularization scheme to achieve temporal smoothness. For example, Chen and Cichocki \cite{chen2005nonnegative} developed a non-negative matrix factorization model with temporal smoothness by constructing a difference term using Toeplitz matrix. To incorporate graph-based side information (e.g., social network), Rao \emph{et al.} \cite{rao2015collaborative} integrated graph Laplacian regularization in matrix factorization. This approach can be applied on time series data by simply adding a chain graph on the temporal dimension. Yokota \emph{et al.} \cite{yokota2016smooth} introduced total variation and quadratic variation into tensor factorization to achieve spatial smoothness. Tan \emph{et al.} \cite{tan2016short} reorganized multivariate traffic time series data as a 4-d (sensor$\times$week$\times$day of week$\times$time of day) tensor to impute missing values. Although this approach does not model temporal smoothness explicitly, the factorization on the 4-d structure is able to learn repeated/reproducible temporal patterns (e.g., daily and weekly). In practice, the regularization scheme are very effective in preserving temporal smoothness. However, on the other hand, these models can only perform interpolation (e.g., imputing missing values at $1\le t\le T$) rather than extrapolation (e.g., predicting values at $t=T+1$).

In order to achieve extrapolation/prediction, we need a generative mechanism on $X$ instead of smoothing. On this track, linear dynamical systems/linear Gaussian state space models have been introduced. For example, Sun \emph{et al.} \cite{sun2014collaborative} presented a dynamic matrix factorization (collaborative Kalman filtering) model for large-scale multivariate time series. Rogers \emph{et al.} \cite{rogers2013multilinear} proposed multilinear dynamical systems (MLDS) by integrating LDS and Tucker decomposition to model tensor time series data. Bahador \emph{et al.} \cite{bahadori2014fast} developed a low-rank tensor learning method to efficiently learn patterns from multivariate spatiotemporal data. Cai \emph{et al.} \cite{cai2015facets} developed a probabilistic temporal tensor decomposition model with higher-order temporal dynamics. Yu \emph{et al.} \cite{yu2016temporal} proposed to impose AR process to regularize the temporal factor matrix. Takeuchi \emph{et al.} \cite{takeuchi2017autoregressive} extended \cite{yu2016temporal} to model not only temporal dynamics but also spatial correlations in tensor data by introducing an additional spatial autoregressive mechanism. Jing \emph{et al.} \cite{jing2018highorder} employed AR process constraints on the core tensor in Tucker decomposition. Essentially, these matrix/tensor factorization-based algorithms are scalable to model large-scale spatiotemporal data and offer a natural solution to deal with the missing data problem. However, in modeling the latent variables and temporal smoothness, these models have to introduce various regularization terms and parameters, which need to be tuned carefully to ensure model accuracy and avoid overfitting. The parameter tuning procedure is computationally very expensive, and more importantly, the procedure has to be done for each particular application/task (i.e., input data set) as there exist no universal/automatic solutions.

In addition to state space-based MF models, another direction is to use Gaussian processes to model the latent factors. Wang \emph{et al.} \cite{wang2005gaussian} introduced Gaussian process dynamical systems (GPDS) as an extension of Gaussian process latent variable models \cite{lawrence2003gaussian}. Temporal dynamics is introduced by a dynamical prior on the latent space $X$. High-order models of GPDM are also developed to better capture temporal dynamics \cite{zhao2016high}. However, GPDM are computationally very expensive for large-scale data set with two kernel matrices of $T\times T$. In addition, the estimation of  hyperparameters is complex kernel functions may end up with local optima. Damianou \emph{et al.} \cite{damianou2011variational} developed a variational inference algorithm for GPDS, and Zhao and Sun \cite{zhao2016variational} took a further step by taking the dependencies among different time series into account with a convolved process. By doing so, the estimation and inference on one time series can also leverage information from other time series. Overall, GPDM are computationally very expensive; MF based-models (collaborative filtering), on the hand, can also achieve information sharing through the low-rank assumption with less cost.
}

\subsection{Bayesian Factorization for Incomplete Matrices/Tensors}

Despite the parameter tuning problem, most of the factorization models above only provide point estimates for imputation/prediction tasks. This becomes a critical concern for real-world applications that are sensitive to uncertainties and risks. Since the introduction of Bayesian Probabilistic Matrix Factorization (BPMF) \cite{salakhutdinov2008bayesian}, Bayesian treatment has been used extensively to address the overfitting and the parameter tuning problems in probabilistic factorization models. {For example,  Luttinen \emph{et al.} \cite{luttinen2012bayesian} developed a Bayesian robust factor analysis (FA) for incomplete matrices, which can better address the overfitting problem in traditional Bayesian PCA \cite{bishop1999bayesian}.} A Bayesian tensor factorization is proposed in \cite{zhao2015bayesianCP}, which can automatically determine the CP rank. Chen \emph{et al.} \cite{chen2019missing} developed an augmented Bayesian tensor factorization model to estimate the posterior distribution of missing values in spatiotemporal traffic data. However, these models essentially focus on the global matrix/tensor factorization without explicitly modeling the local temporal and spatial dependencies in factor matrices. To model temporal smoothness, dynamical factorization models have been developed recently by incorprating a first-order Markovian/state-space assumption. For example, Xiong \emph{et al.} \cite{xiong2010temporalcf} integrated a first-order dynamical structure to characterize temporal dependencies in Bayesian Gaussian tensor factorization. {Luttinen and Ilin \cite{luttinen2009variational}, Adams \emph{et al.} \cite{adams2010incorporating}, and Zhou \emph{et al.} \cite{zhou2012kernelized} imposed Gaussian process priors on the factor matrices in probabilistic factorization, thus allowing factor to incorporate side (spatial/temporal) information using specific kernel structures (e.g., Matern kernel, graph kernel). Although Gaussian process provides a great framework to model spatiotemporal consistency, in the meanwhile it also introduces a critical computational challenge which is estimating the hyperparameters in kernels.}

In this paper, we propose a novel Bayesian Temporal Factorization (BTF) framework that can simultaneously address the regularization parameter tuning problem and the uncertainty estimate problem. BTMF can be considered the Bayesian counterpart of Yu \emph{et al.} \cite{yu2016temporal} by replacing the independent AR assumption on temporal factors with a more flexible VAR process. BTTF, on the other hand, can be considered an extension of the temporal collaborative filtering model by Xiong \emph{et al.} \cite{xiong2010temporalcf} with a VAR prediction mechanism.

\section{Problem Description} \label{sec:problem_description}

We assume a spatiotemporal setting for multidimensional time series data throughout this paper. In general, modern spatiotemporal data sets collected from sensor networks can be organized as matrix and tensor time series. For example, we denote by matrix $Y\in\mathbb{R}^{N\times T}$ a multivariate time series collected from $N$ locations/sensors on $T$ time stamps, with each row $$\boldsymbol{y}_{i}=\left(y_{i,1},y_{i,2},\ldots,y_{i,t-1},y_{i,t},y_{i,t+1},\ldots,y_{i,T}\right)$$
corresponding to the time series collected at location/sensor $i$. As another example, the time-varying origin-destination travel demand can be organized as a third-order time series tensor $\mathcal{Y}\in\mathbb{R}^{M \times N \times T}$ with $M$ origin zones and $N$ destination zones ($M=N$ in most cases), with each time series
$$\boldsymbol{y}_{i,j}=\left(y_{i,j,1},y_{i,j,2},\ldots,y_{i,j,t-1},y_{i,j,t},y_{i,j,t+1},\ldots,y_{i,j,T}\right)$$ showing the number of trips from $i$ to $j$ over time. Given the dimension/number of attributes collected from the underlying system, this formulation can be further extended to even higher-order tensors.

\begin{figure}[ht!]
\centering
\subfigure[Matrix time series]{
    \centering
    \includegraphics[scale=0.76]{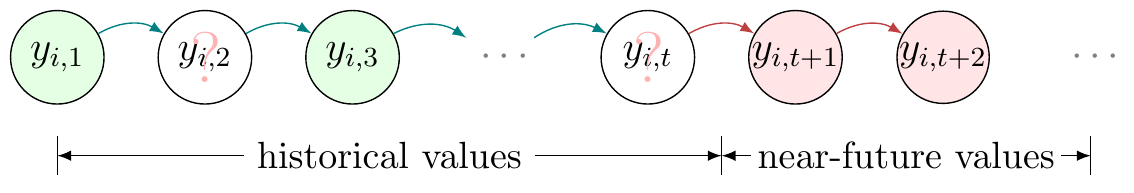}
}
\subfigure[Tensor time series]{
    \centering
    \includegraphics[scale=0.76]{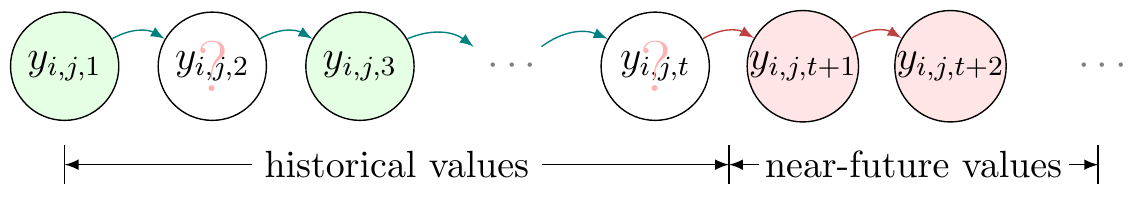}
}
\caption{Illustration of matrix/tensor time series and the prediction problem in the presence of missing values (green: observed data; white: missing data; red: prediction).}
\label{fig:graphical_multivariate_time_series}
\end{figure}

As mentioned above, making accurate predictions on incomplete time series is very challenging, while missing data problem is almost inevitable in real-world applications. Fig.~\ref{fig:graphical_multivariate_time_series} illustrates the prediction problem for incomplete time series data. Here we use $(i,t)\in\Omega$ and $(i,j,t)\in\Omega$ to index the observed entries in matrix $Y$ and tensor $\mathcal{Y}$, respectively.

\section{Bayesian Temporal Matrix Factorization} \label{sec:btmf_model}

\subsection{Model Specification}

Based on the idea of TRMF \cite{yu2016temporal}, here we develop the BTMF framework by employing a Gaussian VAR process to model the temporal factor matrix. Given a partially observed matrix $Y\in\mathbb{R}^{N \times T}$ in a spatiotemporal setting, one can factorize it into a spatial factor matrix $W\in\mathbb{R}^{R \times N}$ and a temporal factor matrix $X\in\mathbb{R}^{R \times T}$ following general matrix factorization model:
\begin{equation}
Y\approx W^{\top}X ,
\label{btmf_equation1}
\end{equation}
and element-wise, we have
\begin{equation}
y_{i,t} = \boldsymbol{w}_{i}^\top\boldsymbol{x}_{t} +\epsilon_{i,t}, \quad \forall (i,t),
\label{btmf_equation2}
\end{equation}
where vectors $\boldsymbol{w}_{i}$ and $\boldsymbol{x}_{t}$ refer to the $i$-th column of $W$ and the $t$-th column of $X$, respectively, and $\epsilon_{i,t}$ is a zero-mean noise term.

\begin{figure}[!ht]
\centering
\includegraphics[scale=0.63]{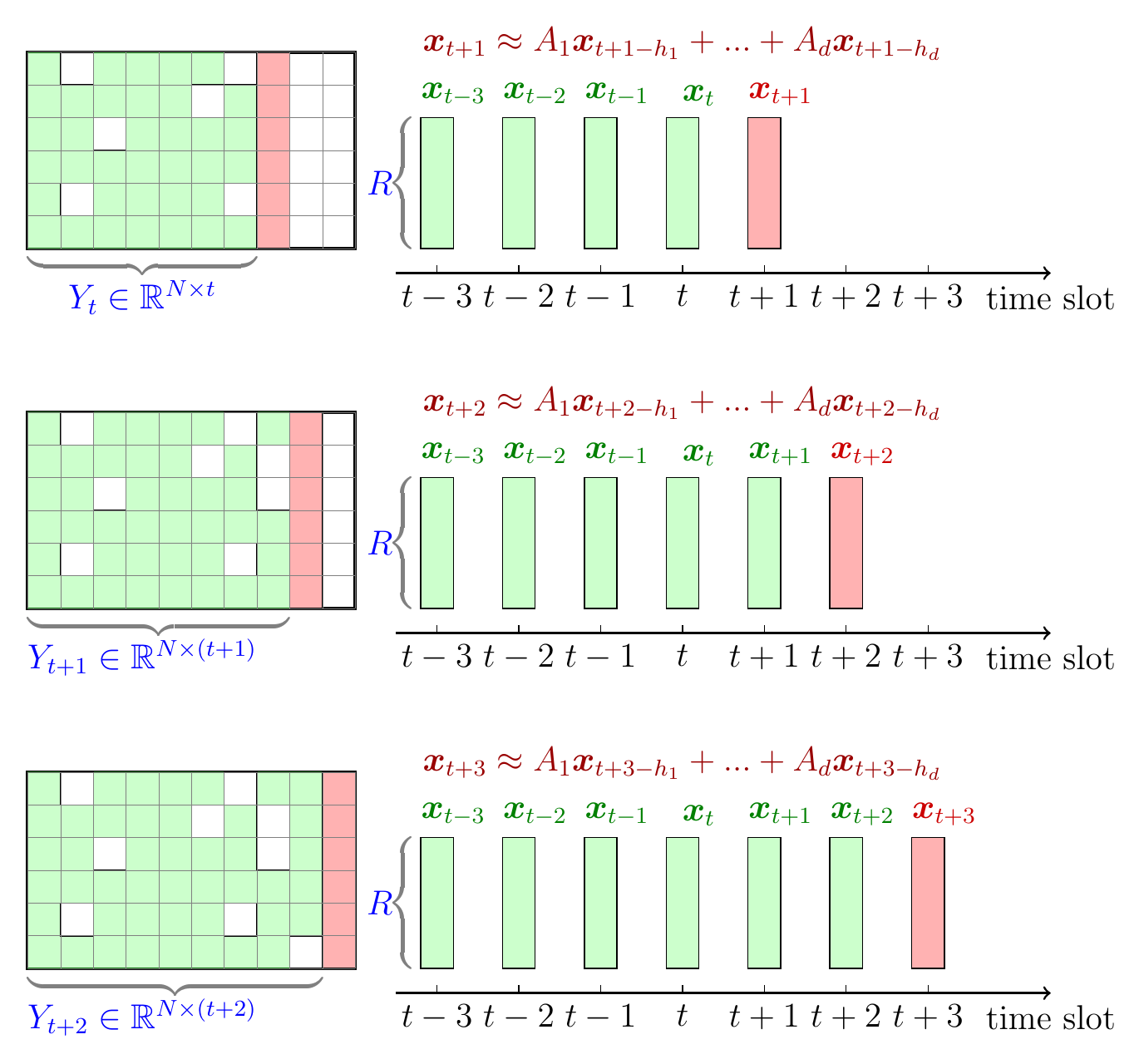}
\caption{A graphical illustration of the rolling prediction scheme using temporal matrix factorization (green: observed data; white: missing data; red: prediction).}
\label{rolling_prediction}
\end{figure}

The standard matrix factorization model is a good approach to deal with the missing data problem; however, it cannot capture the temporal dependencies among different columns in $X$, which are critical in modeling time series data. To characterize the temporal dependencies, an AR regularizer on $X$ is introduced in TRMF \cite{yu2016temporal}:
\begin{equation} \label{equ:AR}
\boldsymbol{x}_{t+1}=\sum\nolimits_{k=1}^{d}\boldsymbol{\theta}_{k}\circledast\boldsymbol{x}_{t+1-h_k}+\boldsymbol{u}_t,
\end{equation}
where $\mathcal{L}=\left\{h_1,\ldots,h_k,\ldots,h_d\right\}$ is a time lag set ($d$ is the order of this AR model), $\boldsymbol{\theta}_{k}$ is a $R\times 1$ coefficient vector, the symbol $\circledast$ denotes the element-wise Hadamard product, and $\boldsymbol{u}_t$ is a Gaussian noise vector. The formulation assumes the temporal dynamics of factors are independent, and thus TRMF can be efficiently estimated using Graph Regularized Alternating Least Squares (GRALS) \cite{rao2015collaborative}. Given observed $Y$ and a trained model, one can first predict $\hat{\boldsymbol{x}}_{t+1}$ on the latent temporal factor matrix $X$ and then estimate time series data at $t+1$ with $y_{i,t+1}\approx \boldsymbol{w}_{i}^\top \hat{\boldsymbol{x}}_{t+1}$. Fig.~\ref{rolling_prediction} illustrates a one-step rolling prediction scheme with real-time data. Therefore, by performing prediction on $X$ instead of on $Y$, TRMF offers a scalable ($R\ll N$) and flexible scheme to model multivariate time series data.

\begin{figure}[!ht]
\centering
\includegraphics[scale=0.73]{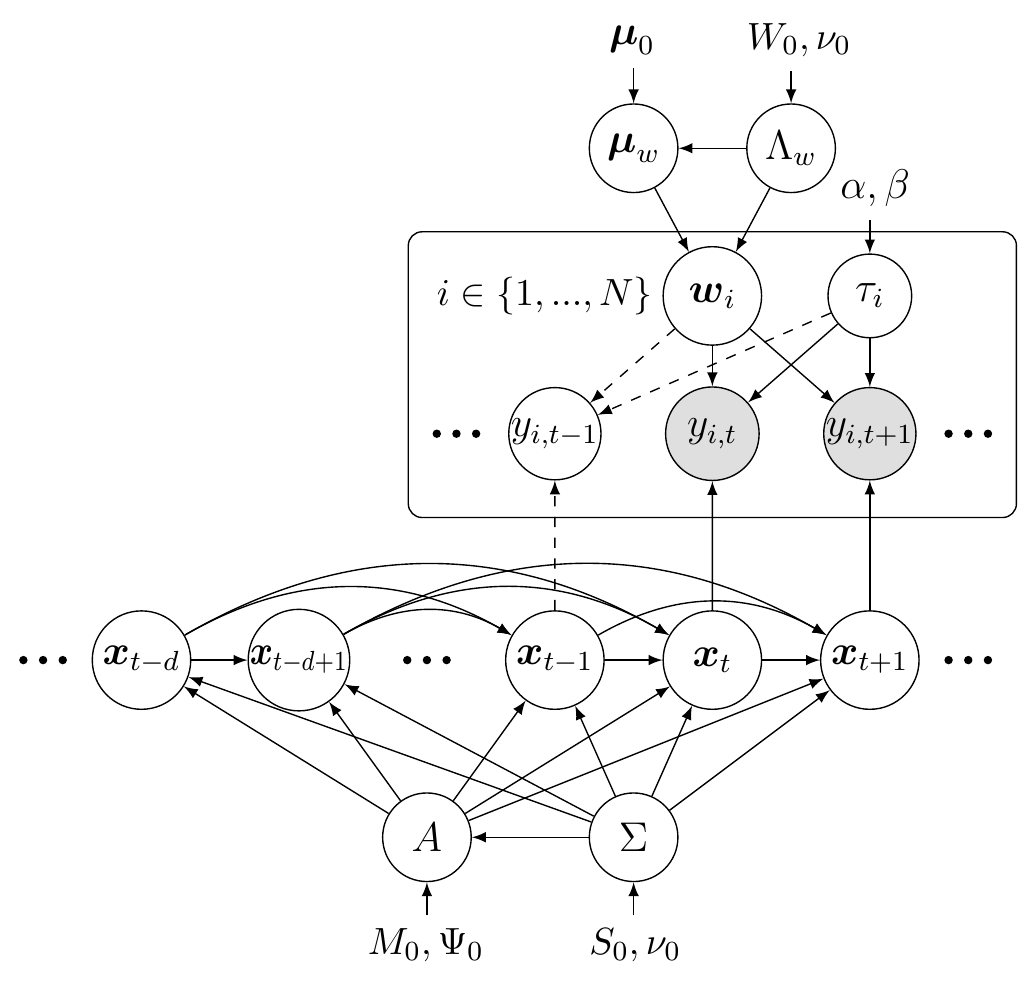}
\caption{An overview graphical model of BTMF (time lag set: $\left\{1,2,\ldots,d\right\}$). The shaded nodes ($y_{i,t}$) are the observed data in $\Omega$.}
\label{btmf_net}
\end{figure}

{
However, TRMF has two major limitations in practice. First, although the independent factor assumption in Eq.~\eqref{equ:AR} greatly reduces the number of parameters, the complex temporal dynamics, causal relationships and covariance structure are essentially ignored. Second, as mentioned before, TRMF requires careful tuning of multiple regularization parameters. The model may end up with overfitting if these regularization parameters are not tuned appropriately. Despite existing parameter tuning solutions (e.g., cross-validation), it is still computationally very expensive to tune a model with many regularization parameters by grid search.

In BTMF, we adopt the VAR process to characterize dynamic dependencies in $X$ instead of AR:
\begin{equation} \label{equ:VAR}
    \boldsymbol{x}_{t+1}=\sum\nolimits_{k=1}^{d}A_{k}\boldsymbol{x}_{t+1-h_k}+\boldsymbol{u}_t,
\end{equation}
where each coefficient matrix $A_k$ ($k\in\left\{1,...,d\right\}$) is of size $R\times R$, and $\boldsymbol{u}_t$ is i.i.d. vector from $\mathcal{N}\left(\boldsymbol{0},\Sigma\right)$. For simplicity, we introduce matrix $A\in \mathbb{R}^{(R d) \times R}$ and vector $\boldsymbol{v}_{t+1}\in \mathbb{R}^{(R d) \times 1}$ as:
\begin{equation*}
A=\left[A_{1}, \ldots, A_{d}\right]^{\top} ,\quad \boldsymbol{v}_{t+1}=\left[\begin{array}{c}{\boldsymbol{x}_{t+1-h_1}} \\ {\vdots} \\ {\boldsymbol{x}_{t+1-h_d}}\end{array}\right],
\end{equation*}
to summarize all coefficient matrices and correlated vectors. Therefore, we have $\boldsymbol{x}_{t+1}=A^\top \boldsymbol{v}_{t+1}+\boldsymbol{u}_{t}$.

Following the principle framework of existing Bayesian probabilistic matrix/tensor factorization models (e.g., BPMF in \cite{salakhutdinov2008bayesian} and BPTF in \cite{xiong2010temporalcf}), we build the BTMF framework following the graphical representation in Fig.~\ref{btmf_net}. Note that this model is entirely built on observed data in $\Omega$ and thus it can be trained on data sets with missing values. We next introduce each component in this graphical model in detail.
}

We assume that observed entries are independent in $Y$ and each entry follows i.i.d. Gaussian distribution with precision $\tau_i$ ($i\in\{1,\ldots,N\}$):
\begin{equation}
y_{i,t}\sim\mathcal{N}\left(\boldsymbol{w}_i^\top\boldsymbol{x}_t,\tau_i^{-1}\right),\quad \left(i,t\right)\in\Omega,
\label{btmf_equation3}
\end{equation}
where $\tau_i$ is a spatially-varying precision characterizing the noise level of time series $\boldsymbol{y}_{i}$. { This assumption corresponds to that the error term $\epsilon_{i,t}$ in Eq.~\eqref{btmf_equation2} follows i.i.d. Gaussian $\epsilon_{i,t}\sim \mathcal{N}\left(0,\tau_i^{-1}\right)$ and $\tau_i\neq \tau_j$ as in factor analysis (FA) \cite{luttinen2012bayesian}. TRMF and other probabilistic matrix factorization models essentially follow the PCA scheme, assuming the noise to be isotropic (i.e., $\tau_i=\tau_j=\tau$). BPMF  \cite{salakhutdinov2008bayesian} and BPTF \cite{xiong2010temporalcf} follow the isotropic noise assumption, which is reasonable since the scale of the data (e.g., rating from 0 to 5) is known in advance. However, for multivariate time series, we may expect different series to have a different level of noise. Therefore, we relax the isotropic noise assumption and make $\tau_i$ a spatially-varying parameter in the form of FA.

}

On the spatial factors, we use a simple Gaussian factor matrix without imposing any dependencies explicitly. The prior of vector $\boldsymbol{w}_{i}$ (i.e., $i$-th column of $W$) is a multivariate Gaussian distribution:
\begin{equation}
\boldsymbol{w}_i\sim\mathcal{N}\left(\boldsymbol{\mu}_{w},\Lambda_w^{-1}\right),
\end{equation}
and we place a conjugate Gaussian-Wishart prior on the mean vector and the precision matrix:
\begin{equation} \label{equ:gaussian_wishart}
\boldsymbol{\mu}_w | \Lambda_w \sim\mathcal{N}\left(\boldsymbol{\mu}_0,(\beta_0\Lambda_w)^{-1}\right),\Lambda_w\sim\mathcal{W}\left(W_0,\nu_0\right),
\end{equation}
where $\boldsymbol{\mu}_0\in \mathbb{R}^{R}$ is a mean vector, and $\mathcal{W}\left(W_0,\nu_0\right)$ is a Wishart distribution with a $R$-by-$R$ scale matrix $W_0$ and $\nu_0$ degrees of freedom.

In modeling the temporal factor matrix $X$, we re-write the VAR process as:
\begin{equation}
\begin{aligned}
\boldsymbol{x}_{t}&\sim\begin{cases}
\mathcal{N}\left(\boldsymbol{0},I_R\right),&\text{if $t\in\left\{1,2,\ldots,h_d\right\}$}, \\
\mathcal{N}\left(A^\top \boldsymbol{v}_{t},\Sigma\right),&\text{otherwise}.\\
\end{cases}\\
\end{aligned}
\label{btmf_equation5}
\end{equation}

Since the mean vector is defined by VAR, we place the conjugate matrix normal inverse Wishart (MNIW) prior on the coefficient matrix $A$ and the covariance matrix $\Sigma$ as follows,
\begin{equation}
\begin{aligned}
A\sim\mathcal{MN}_{(Rd)\times R}\left(M_0,\Psi_0,\Sigma\right),\quad
\Sigma \sim\mathcal{IW}\left(S_0,\nu_0\right), \\
\end{aligned}
\end{equation}
where the probability density function for the $Rd$-by-$R$ random matrix $A$ has the form:
\begin{equation}
\begin{aligned}
&p\left(A\mid M_0,\Psi_0,\Sigma\right)=\left(2\pi\right)^{-R^2d/2}\left|\Psi_0\right|^{-R/2}\left|\Sigma\right|^{-Rd/2} \\
& \times\exp\left(-\frac{1}{2}\text{tr}\left[\Sigma^{-1}\left(A-M_0\right)^{\top}\Psi_{0}^{-1}\left(A-M_0\right)\right]\right), \\
\end{aligned}
\label{mnpdf}
\end{equation}
with $\Psi_0\in\mathbb{R}^{(Rd)\times (Rd)}$ and $\Sigma\in\mathbb{R}^{R\times R}$ as covariance matrices.

For the only remaining parameter $\tau_i$, we place a Gamma prior  $\tau_i\sim\text{Gamma}\left(\alpha,\beta\right)$, where $\alpha$ and $\beta$ are the shape and rate parameters, respectively.

The above specifies the full generative process of BTMF. Several parameters are introduced to define the prior distributions, including $\boldsymbol{\mu}_{0}$, $W_0$, $\nu_0$, $\beta_0$, $\alpha$, $\beta$, $M_0$, $\Psi_0$, and $S_0$. These parameters need to be specified in advance. {In the implementation, we use the same non-informative priors as in \cite{salakhutdinov2008bayesian,xiong2010temporalcf,zhao2015bayesianCP} by setting $\beta_0=1$, $\nu_0=R$, $\boldsymbol{\mu}_0=\boldsymbol{0}$ as a zero vector, $W_0=I_R$, $S_0=I_R$, $\Psi_0=I_{Rd}$, $M_0=0$ as a zero matrix, and $\alpha=\beta=10^{-6}$. The weak specification of these parameters has little impact on the final results, as the large training data (i.e., likelihood) will play a much more important role in defining the posterior.}

{

It should be noted that the proposed model can be considered a special case of dynamic factor model (DFM)  \cite{koop2010bayesian,stock2016dynamic}:
\begin{equation} \label{equ:DFM}
    \begin{aligned}
    y_{i,t} &= \boldsymbol{w}_{i}^\top\boldsymbol{x}_{t} +\epsilon_{i,t},\\
    \boldsymbol{x}_{t+1}&= \sum\nolimits_{k=1}^{d}A_{k}\boldsymbol{x}_{t+1-h_k}+\boldsymbol{u}_t,
    \end{aligned}
\end{equation}
where we assume $\boldsymbol{\epsilon}_t$ to be serially uncorrelated (i.e., $\boldsymbol{\epsilon}_t \sim \text{ i.i.d. } \mathcal{N}\left(0,\Phi\right)$ with $\Phi$ being a diagonal matrix filled by $\tau_i$).

The DFM in Eq.~\eqref{equ:DFM} has been extensively studied in economics and statistics literature \cite{koop2010bayesian,stock2016dynamic}. However, most studies focuses on the identification problem (for better interpretation) and serial correlation. Two sets of parameters $\{W,A,\{\tau\}_{i=1}^N,\Sigma\}$ and $\{WP^{\top},PA,\{\tau\}_{i=1}^N,P\Sigma P^{\top}\}$ are equivalent for any $R\times R$ orthogonal matrix $P$. For example, to ensure identification, Aguilar and West \cite{aguilar2000bayesian} proposed to restrict $W$ to be a block lower triangular matrix with each row having a separate and fixed prior. This assumption, however, restricts the applicability of the model, since the priors are essentially unknown but important for large data sets.

For the time series prediction tasks, we would like to highlight that we are more interested in the accuracy and uncertainty of the prediction results than the identification of parameters. Therefore, the proposed BTMF simply follows the general BPMF \cite{salakhutdinov2008bayesian,xiong2010temporalcf} framework by placing a flexible conjugate Gaussian-Wishart prior on the mean vector and precision matrix of $\boldsymbol{w}_i$ as in Eq.~\eqref{equ:gaussian_wishart}, without imposing any identification constraints.

}

\subsection{Model Inference}\label{sec:btmf_inference}

Here we rely on the MCMC technique to perform Bayesian inference. Specifically, we introduce a Gibbs sampling algorithm by deriving the full conditional distributions for all parameters and hyperparameters. Thanks to the use of conjugate priors in Fig.~\ref{btmf_net}, we can write down all the conditional distributions analytically. Below we summarize the Gibbs sampling procedure.

\noindent\textbf{{Sampling $\left(\boldsymbol{\mu}_{w},\Lambda_{w}\right)$}}. The conditional distribution is given by a Gaussian-Wishart:
\begin{equation*}
p\left(\boldsymbol{\mu}_{w},\Lambda_{w}|-\right)=\mathcal{N}(\boldsymbol{\mu}_{w}^{*},\left(\left(\beta_0+N\right)\Lambda_{w}\right)^{-1})\times \mathcal{W}\left(W_{w}^{*},\nu_{w}^{*}\right),
\end{equation*} where
\begin{multline*}
\boldsymbol{\mu}_{w}^{*}=\frac{1}{\beta_0+N}\left(\beta_0\boldsymbol{\mu}_0+N\bar{\boldsymbol{w}}\right),\quad\nu_{w}^{*}=\nu_0+N, \\
\left(W_{w}^{*}\right)^{-1}=W_0^{-1}+NS_{w}+\frac{\beta_0N}{\beta_0+N}\left(\bar{\boldsymbol{w}}-\boldsymbol{\mu}_0\right)\left(\bar{\boldsymbol{w}}-\boldsymbol{\mu}_0\right)^\top,\\
\bar{\boldsymbol{w}}=\frac{1}{N}\sum\nolimits_{i=1}^{N}\boldsymbol{w}_{i}, \quad
S_{w}=\frac{1}{N}\sum\nolimits_{i=1}^{N}\left(\boldsymbol{w}_{i}-\bar{\boldsymbol{w}}\right)\left(\boldsymbol{w}_{i}-\bar{\boldsymbol{w}}\right)^\top.
\end{multline*}

\noindent\textbf{{Sampling $\left(A,\Sigma\right)$}}. Given the MNIW prior, the corresponding conditional distribution is
\begin{equation}
\begin{aligned}
&p\left(A,\Sigma|-\right)
=\mathcal{MN}\left(M^{*},\Psi^{*},\Sigma\right)\times\mathcal{IW}\left(S^{*},\nu^{*}\right),
\end{aligned}
\label{mniw_posterior}
\end{equation}
and the parameters are given by:
\begin{equation*}
\begin{aligned}
&\Psi^{*}=\left(\Psi_{0}^{-1}+Q^{\top} Q\right)^{-1},\\
&M^{*}=\Psi^{*}\left(\Psi_{0}^{-1} M_{0}+Q^{\top} Z\right), \\
&S^{*}=S_{0}+Z^\top Z+M_0^\top\Psi_0^{-1}M_0-\left(M^{*}\right)^\top\left(\Psi^{*}\right)^{-1}M^{*}, \\
&\nu^{*}=\nu_{0}+T-h_d,
\end{aligned}
\end{equation*}
where the matrices $Z\in\mathbb{R}^{(T-h_d) \times R}$ and $Q\in \mathbb{R}^{(T-h_d) \times(R d)}$ are defined as:
\begin{equation*}
Z=\left[\begin{array}{c}{\boldsymbol{x}_{h_d+1}^{\top}} \\ {\vdots} \\ {\boldsymbol{x}_{T}^{\top}}\end{array}\right],\quad Q=\left[\begin{array}{c}{\boldsymbol{v}_{h_d+1}^{\top}} \\ {\vdots} \\ {\boldsymbol{v}_{T}^{\top}}\end{array}\right].
\end{equation*}

\noindent\textbf{{Sampling spatial factor $\boldsymbol{w}_{i}$}}. The conditional posterior distribution $p\left(\boldsymbol{w}_{i}\mid\boldsymbol{y}_{i},X,\tau_i,\boldsymbol{\mu}_{w},\Lambda_{w}\right)$ is a Gaussian distribution. Thus, we can sample $\boldsymbol{w}_{i}|-\sim\mathcal{N}(\boldsymbol{\mu}_{i}^{*},\left(\Lambda_{i}^{*}\right)^{-1})$ with
\begin{equation} \label{equ:sample_w}
\begin{aligned}
\Lambda_{i}^{*}&=\tau_i\sum\nolimits_{t:(i,t)\in\Omega}\boldsymbol{x}_{t}\boldsymbol{x}_{t}^\top+\Lambda_{w}, \\
\boldsymbol{\mu}_{i}^{*}&=\left(\Lambda_{i}^{*}\right)^{-1}\left(\tau_i\sum\nolimits_{t:(i,t)\in\Omega}\boldsymbol{x}_{t}y_{i,t}+\Lambda_{w}\boldsymbol{\mu}_{w}\right).
\end{aligned}
\end{equation}

\noindent\textbf{{Sampling temporal factor $\boldsymbol{x}_{t}$}}. Following \cite{xiong2010temporalcf}, we perform a single-site update on the latent variable $\boldsymbol{x}_{t}$. Given the VAR process, the conditional distribution of $\boldsymbol{x}_{t}$ is also a Gaussian. However, for a particular time lag set, we need to define different updating rules for $1\le t\le T-h_1$ and $T-h_1< t\le T$. Overall, the conditional distribution can be written as $p\left(\boldsymbol{x}_{t}|-\right)=\mathcal{N}(\boldsymbol{\mu}_{t}^{*},\Sigma_{t}^{*})$ with
\begin{equation}
\begin{aligned}
{\Sigma}_{t}^{*}&=\left(\sum\nolimits_{i:(i,t)\in\Omega}\tau_i\boldsymbol{w}_{i}\boldsymbol{w}_{i}^\top+M_{t}+P_{t}\right)^{-1}, \\
{\boldsymbol{\mu}}_{t}^{*}&=\Sigma_{t}^{*}\left(\sum\nolimits_{i:(i,t)\in\Omega}\tau_i\boldsymbol{w}_{i}y_{i,t}+N_{t}+Q_{t}\right),
\end{aligned}
\label{btmf_equation11}
\end{equation}
where $M_t$ and $N_t$ are two auxiliary variables. In general cases where $1\le t\le T-h_1$, we define $M_t$ and $N_t$ as:
\begin{equation*}
\begin{aligned}
M_t&=\sum\nolimits_{k=1,h_d<t+h_k\leq T}^dA_k^\top\Sigma^{-1} A_k, \\
N_t&=\sum\nolimits_{k=1,h_d<t+h_k\leq T}^dA_k^\top\Sigma^{-1}\boldsymbol{\psi}_{t+h_k},\\
\boldsymbol{\psi}_{t+h_k}&=\boldsymbol{x}_{t+h_k}-\sum\nolimits_{l=1,l\neq k}^{d}A_l\boldsymbol{x}_{t+h_k-h_l}.
\end{aligned}
\label{btmf_equation12}
\end{equation*}
Otherwise, we define $M_t=0$ and $N_t=0$.

The variables $P_{t}$ and $Q_{t}$ in Eq.~\eqref{btmf_equation11} are given by:
\begin{equation*}
\begin{aligned}
P_t&=\begin{cases}
I_R, & \text{if $t\in\left\{1,2,\ldots,h_d\right\}$}, \\
\Sigma^{-1}, & \text{otherwise},
\end{cases} \\
Q_t&=\begin{cases}
\boldsymbol{0}, & \text{if $t\in\left\{1,2,\ldots,h_d\right\}$}, \\
\Sigma^{-1}\sum\nolimits_{l=1}^{d}A_l\boldsymbol{x}_{t-h_l}, & \text{otherwise}.
\end{cases} \\
\end{aligned}
\label{btmf_equation14}
\end{equation*}

{
Noted that the single-site update for $\boldsymbol{x}_t$ is easy to implement but it may give slow mixing for large $T$. Our numerical experiment shows this simple approach is still very efficient. One can also apply the forward-filtering backward-sampling algorithm/Kalman filter for state-space models with higher computational cost (see e.g., \cite{west1997bayesian}) so that updating $\boldsymbol{x}_{t}$ may benefit from faster mixing.
}

\noindent\textbf{{Sampling precision $\tau_i$}}. Given the conjugate Gamma prior, the conditional distribution of $\tau_i$ is also a Gamma distribution. Thus, we sample $\tau_i|-\sim \text{Gamma}\left(\alpha_i^{*},\beta_i^{*}\right)$ with
\begin{equation}\label{equ:tau}
\begin{aligned}
\alpha_i^{*}&=\frac{1}{2}\sum\nolimits_{t:(i,t)\in\Omega}\mathbbm{1}(y_{it})+\alpha, \\
\beta_i^{*}&=\frac{1}{2}\sum\nolimits_{t:(i,t)\in\Omega}(y_{it}-\boldsymbol{w}_{i}^\top\boldsymbol{x}_{t})^2+\beta,
\end{aligned}
\end{equation}
where the operator $\mathbbm{1}(\cdot)$ denotes the indicator function that holds $\mathbbm{1}(y_{it})=1$ if $y_{it}$ is observed, and $\mathbbm{1}(y_{it})=0$ otherwise.

\subsection{Model Implementation}\label{sec:btmf_implementation}

\noindent\textbf{BTMF imputation}. Based on the aforementioned sampling processes, we summarize the MCMC inference algorithm to impute missing values in the partially observed matrix time series data as Algorithm~\ref{btmf_algorithm}. In training the model, we first run the MCMC algorithm for $m_1$ iterations as burn-in and then take samples from the following $m_2$ iterations for estimation. Note that one can keep all the $m_2$ samples to compute not only the mean as a point estimate but also the confidence interval and the variance for risk-sensitive applications.

\begin{algorithm}
 \caption{BTMF imputation}
 \label{btmf_algorithm}
\begin{algorithmic}[1]
 \renewcommand{\algorithmicrequire}{\textbf{Input:}}
 \renewcommand{\algorithmicensure}{\textbf{Output:}}
 \REQUIRE Data matrix $Y\in\mathbb{R}^{N\times T}$, $\Omega$ as the set of observed entries in $Y$, $\mathcal{L}=\left\{h_1,h_2,\ldots,h_d\right\}$ as the time lag set, number of burn-in iterations $m_1$, and number of samples used in estimation $m_2$.
 Initialization of factor matrices $\left\{W,X\right\}$ and VAR coefficient matrix  $A\in\mathbb{R}^{Rd\times R}$.
 \ENSURE Estimated matrix $\hat{Y}\in\mathbb{R}^{N\times T}$.
  \FOR {$\text{iter.} = 1$ to $m_1+m_2$}
  \STATE Draw hyperparameters $\left\{\boldsymbol{\mu}_{w},\Lambda_{w}\right\}$.
  \FOR {$i=1$ to $N$ (can be in parallel)}
  \STATE Draw $\boldsymbol{w}_{i}\sim\mathcal{N}(\boldsymbol{\mu}_{w}^{*},(\Lambda_{w}^{*})^{-1})$.
  \ENDFOR
  \STATE Draw $\Sigma\sim\mathcal{IW}\left(S^{*},\nu^{*}\right)$ and $A\sim\mathcal{MN}\left(M^{*},\Psi^{*},\Sigma\right)$.
  \FOR {$t=1$ to $T$}
  \STATE Draw $\boldsymbol{x}_{t}\sim\mathcal{N}(\boldsymbol{\mu}_{t}^{*},\Sigma_{t}^{*})$.
  \ENDFOR
  \FOR {$i=1$ to $N$}
  \STATE Draw precision $\tau_i\sim\text{Gamma}(\alpha_{i}^{*},\beta_{i}^{*})$.
  \ENDFOR
  \IF {$\text{iter.} > m_1$}
  \STATE Compute $\tilde{Y}=W^{\top}X$. Collect sample $\tilde{Y}$.
  \ENDIF
  \ENDFOR
\RETURN $\hat{Y}$ as the average of the $m_2$ samples of $\tilde{Y}$.
\end{algorithmic}
\end{algorithm}

{

\noindent\textbf{BTMF forecasting}. To better support real-time prediction, we build an efficient multi-step rolling prediction scheme based on Algorithm~\ref{btmf_algorithm}. The key idea is to update model parameters when new data comes instead of performing full retraining as in TRMF. In doing so, we only consider the future  latent factor $X$ as new parameters to be sampled/updated over time as in \cite{Gultekin2019online}.

Assume that we have historical data $Y\in\mathbb{R}^{N\times T}$. Denote by $\delta$ the window length of multi-step predictions and $S$ the total rolling windows. To make prediction with BTMF for $s=1$ (i.e., initial prediction), we gather $m_2$ samples which consist of $\{X^{(\ell)},W^{(\ell)},A^{(\ell)},\Sigma^{(\ell)}\}_{\ell=1}^{m_2}$ as shown in Algorithm~\ref{btmf_algorithm} and draw predicted temporal factors as follows,
\begin{equation}\label{sample_temp_fact}
\begin{aligned}
\boldsymbol{x}_{T+1}&\sim \mathcal{N}\left(\sum\nolimits_{k}A_{k}^{(\ell)}\boldsymbol{x}_{T+1-h_k}^{(\ell)},\Sigma^{(\ell)}\right), \\
\vdots& \\
\boldsymbol{x}_{T+\delta}&\sim \mathcal{N}\left(\sum\nolimits_{k}A_{k}^{(\ell)}\boldsymbol{x}_{T+\delta-h_k}^{(\ell)},\Sigma^{(\ell)}\right), \\
\end{aligned}
\end{equation}
where we have $X_{1}^{(\ell)}=(\boldsymbol{x}_{T+1},\cdots,\boldsymbol{x}_{T+\delta})\in\mathbb{R}^{R\times\delta}$, and the average of $W^{(\ell)\top} X_{1}^{(\ell)},\ell=1,\ldots,m_2$ is considered as the predicted values. We then combine $X^{(\ell)}$ and $X_{1}^{(\ell)}$ to perform prediction on the next rolling window (i.e., $s=2$).

Given that $Y$ contains substantial missing and corruption, the estimation of $X_1^{(\ell)}$ ($s=1$) may have large error and the rolling scheme will spread the error to the estimation of $X_2^{(\ell)}$ ($s=2$). To alleviate this effect, we use the last $\gamma\cdot\delta$ ($\gamma$ is a positive integer) columns of $Y_{T+\delta(s-1)}\in\mathbb{R}^{N\times (T+\delta(s-1))}$ as the incremental data matrix
\begin{equation} \notag
    Y_{s} = (\boldsymbol{y}_{T+\delta(s-\gamma-1)+1},\boldsymbol{y}_{T+\delta(s-\gamma-1)+2},\ldots,\boldsymbol{y}_{T+\delta(s-1)})
\end{equation}
of size $N\times (\gamma\cdot\delta)$ to re-update the corresponding $\boldsymbol{x}_{t}$s. Thus, $\gamma=1$ corresponds to the case where we only use the most recent values (i.e., streaming data) to update $\boldsymbol{x}$. By applying the samples of $\{X^{(\ell)},W^{(\ell)},\boldsymbol{\tau}^{(\ell)},A^{(\ell)},\Sigma^{(\ell)}\}_{\ell=1}^{m_2}$, we first draw temporal factors $\boldsymbol{x}_{t}\sim\mathcal{N}\left(\boldsymbol{\mu}_{t}^{*},\Sigma_{t}^{*}\right)$ as defined in Eq.~\eqref{btmf_equation11} for time $T+\delta(s-\gamma-1)+1$ to $T+\delta(s-1)$. Then, we collect the samples of predicted temporal factors as in Eq.~\eqref{sample_temp_fact} and make predictions for time $T+\delta(s-1)+1$ to $T+\delta s$ accordingly. We summarize the implementation of BTMF forecasting in Algorithm~\ref{btmf_forecasting}.

\begin{algorithm}
 \caption{BTMF forecasting}
 \label{btmf_forecasting}
\begin{algorithmic}[1]
 \renewcommand{\algorithmicrequire}{\textbf{Input:}}
 \renewcommand{\algorithmicensure}{\textbf{Output:}}
 \REQUIRE $\mathcal{L}=\left\{h_1,h_2,\ldots,h_d\right\}$ as the time lag set, and number of samples used in estimation $m_2$.
 \ENSURE Predicted matrix $\hat{Y}\in\mathbb{R}^{N\times (\delta\cdot S)}$.
 \\ Initialize $\beta_0=1$, $\nu_0=R$. Set $\gamma=10$.
 \STATE Train BTMF (Algorithm~\ref{btmf_algorithm}) on data $Y\in\mathbb{R}^{N\times T}$, collect $m$ samples of $W,X,A,\boldsymbol{\tau},\Sigma$, and make predictions on temporal factors as shown in Eq.~\eqref{sample_temp_fact}.
 \FOR {$s=2$ to $S$}
  \STATE Gather the incremental data $Y_{s}\in\mathbb{R}^{N\times(\gamma\cdot\delta)}$.
  \FOR {$\ell = 1$ to $m_2$}
  \STATE Draw $\gamma\cdot\delta$ temporal factors.
  \FOR {$\kappa=1$ to $\delta$ (make prediction)}
  \STATE $t_0=T+\delta(s-1)+\kappa$, draw $\boldsymbol{x}_{t_0}\sim\mathcal{N}(\boldsymbol{\mu}_{t_0}^{*},\Sigma_{t_0}^{*})$.
  \ENDFOR
  \STATE Compute $\tilde{Y}_{s}^{(\ell)}=(W^{(\ell)})^{\top}X_{s}^{(\ell)}$.
  \ENDFOR
  \STATE Collect sample $\tilde{Y}_{s}=\{\tilde{Y}_{s}^{(1)},\ldots,\tilde{Y}_{s}^{(m_2)}\}$.
 \ENDFOR
\RETURN $\hat{Y}$ as the average of the $m_2$ samples of $\tilde{Y}$.
\end{algorithmic}
\end{algorithm}

Since the Bayesian model has been trained using all available data, the MCMC algorithm for updating temporal factors is expected to converge very fast in a few iterations and the $m_2$ samples can be generated very efficiently. Again, we would like to emphasize that we only consider temporal factors $\boldsymbol{x}_t$ as the variable to sample to achieve high efficiency in this rolling prediction application. In the meanwhile, as the updating is only made on $X$, we would still expect significant error accumulation when $\gamma$ is small. To address this issue, one may choose to make a full retrain when the prediction error becomes large. For example, we can following the incremental/global learning scheme in \cite{deng2016latent} to achieve both efficiency and accuracy.
}

\section{Bayesian Temporal Tensor Factorization} \label{sec:bttf_model}

It is straightforward to extend BTMF to model multidimensional (order$>$2) tensor time series. We use a third-order tensor $\mathcal{Y}\in\mathbb{R}^{M \times N \times T}$ as an example throughout the section.

\subsection{Model Specification}

\begin{figure}[ht!]
\centering
\includegraphics[scale=0.8]{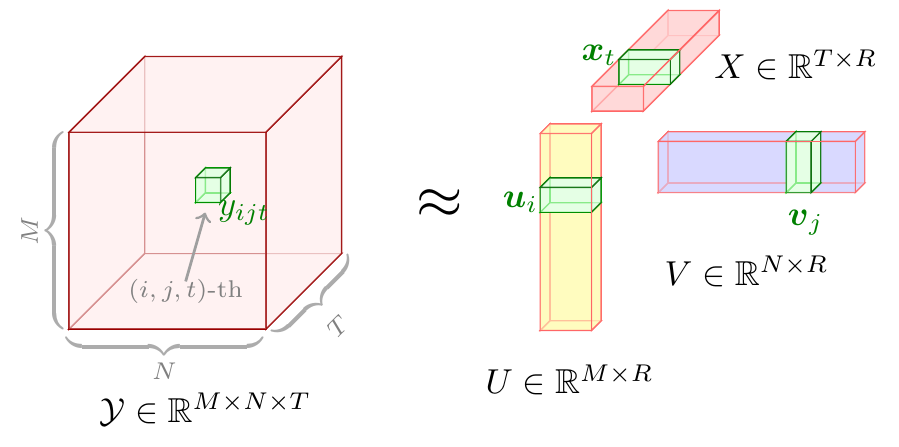}
\caption{A graphical illustration of CP factorization.}
\label{CP_factorization}
\end{figure}

To model multidimensional data, we employ the CANDECOMP/PARAFAC (CP) decomposition \cite{kolda2009tensor}, which approximates $\mathcal{Y}$ by the sum of $R$ rank-one tensors:
\begin{equation}
\mathcal{Y}\approx\sum\nolimits_{r=1}^{R}\boldsymbol{u}_{r}\circ\boldsymbol{v}_{r}\circ\boldsymbol{x}_{r},
\label{bttf_equation1}
\end{equation}
where $\boldsymbol{u}_{r}\in\mathbb{R}^{M}$, $\boldsymbol{v}_{r}\in\mathbb{R}^{N}$, and $\boldsymbol{x}_{r}\in\mathbb{R}^{T}$ are the $r$-th column of factor matrices $U\in\mathbb{R}^{M\times R}$, $V\in\mathbb{R}^{N\times R}$, and $X\in\mathbb{R}^{T\times R}$, respectively (see Fig.~\ref{CP_factorization}). The symbol $\circ$ denotes vector outer product. Essentially, this model can be considered a high-order extension of matrix factorization in Eq.~\eqref{btmf_equation1}. 

{
The CP decomposition provides us a natural way to extend BTMF to tensors by assuming that each element follows a Gaussian distribution:
\begin{equation}
y_{i,j,t}\sim\mathcal{N}\left(\sum\nolimits_{r=1}^{R}u_{ir}v_{jr}x_{tr},\tau_{ij}^{-1}\right), \quad \left(i,j,t\right)\in\Omega,
\label{bttf_equation3}
\end{equation}
where $\tau_{ij}$ is the spatially-varying precision for the $\left(i,j\right)$-th time series. Note that when the tensor is large, sparse, and heterogeneous, it becomes challenging to learn the noise level for each individual time series. In this case, we may assume isotropic noise in the form of probabilistic PCA ($\tau_{ij}=\tau$) for simplicity.
}

Following the same routine as BTMF, we define the generative process of Bayesian Temporal Tensor Factorization (BTTF) as follows,
\begin{equation}
\begin{aligned}
\boldsymbol{u}_i&\sim\mathcal{N}\left(\boldsymbol{\mu}_{u},\Lambda_u^{-1}\right), \\ \boldsymbol{v}_j&\sim\mathcal{N}\left(\boldsymbol{\mu}_{v},\Lambda_v^{-1}\right), \\
\tau &\sim\text{Gamma}\left(\alpha,\beta\right),
\end{aligned}
\label{bttf_equation4}
\end{equation}
and the same VAR model in Eq.~\eqref{btmf_equation5} can be used to model temporal factor matrix $X$, and the prior is defined as:
\begin{equation}
\begin{aligned}
\boldsymbol{x}_{t}&\sim\begin{cases}
\mathcal{N}\left(\boldsymbol{0},I_R\right),&\text{if $t\in\left\{1,2,\ldots,h_d\right\}$}, \\
\mathcal{N}\left(A^\top \boldsymbol{v}_{t},\Sigma\right),&\text{otherwise},\\
\end{cases}\\
\end{aligned}
\label{bttf_equation5}
\end{equation}
where in this setting, the same Gaussian-Wishart priors as in BTMF can be placed on hyperparameters.

In BTTF, we may consider both $U$ and $V$ as spatial factor matrices, while in fact they may characterize any features in which dependencies are not explicitly encoded (e.g., type of sensors in \cite{bahadori2014fast,takeuchi2017autoregressive}). { It should be noted that the BPTF algorithm proposed in \cite{xiong2010temporalcf} is a special case of BTTF. Specifically, BPTF corresponds to the model where $\mathcal{L}=\left\{1\right\}$,  $A_1=I_{R}$ (identity matrix instead of a free parameter matrix), and  noise is isotropic (i.e., $\tau_{ij}=\tau$). }

\subsection{Model Inference}

Regarding posterior inference, the main difference between BTTF and BTMF is the posterior distribution of factor matrices. Specifically, the posterior distribution of $\boldsymbol{x}_{t}$ in BTTF can be written as $p\left(\boldsymbol{x}_{t}|-\right)=\mathcal{N}(\boldsymbol{\mu}_{t}^{*},\Sigma_{t}^{*})$ with
\begin{equation}
\begin{aligned}
{\Sigma}_{t}^{*}&=\left(\sum\nolimits_{i,j:(i,j,t)\in\Omega}\tau_{ij}\boldsymbol{w}_{ij}\boldsymbol{w}_{ij}^\top+M_{t}+P_t\right)^{-1}, \\
{\boldsymbol{\mu}}_{t}^{*}&=\Sigma_{t}^{*}\left(\sum\nolimits_{i,j:(i,j,t)\in\Omega}\tau_{ij}\boldsymbol{w}_{ij}y_{ijt}+N_{t}+Q_{t}\right),
\end{aligned}
\label{bttf_equation11}
\end{equation}
where $\boldsymbol{w}_{ij}=\boldsymbol{u}_{i}\circledast\boldsymbol{v}_{j}\in\mathbb{R}^{R}$, and $\left\{M_t,N_t,P_t,Q_t\right\}$ are defined in the same way as in BTMF (see Eq.~\eqref{btmf_equation11}).

The posterior distribution of $\boldsymbol{u}_{i}$ is $\mathcal{N}(\boldsymbol{u}_{i}|\boldsymbol{\mu}_{i}^{*},\left(\Lambda_{i}^{*}\right)^{-1})$ with
\begin{equation}
    \begin{aligned}
    \Lambda_{i}^{*}&= \sum\nolimits_{j,t :(i,j,t) \in \Omega} \tau_{ij}\boldsymbol{w}_{jt} \boldsymbol{w}_{jt}^{\top}+\Lambda_{u}, \\ \boldsymbol{\mu}_{i}^{*}&=\left(\Lambda_{i}^{*}\right)^{-1}\left(\sum\nolimits_{j,t :(i,j,t) \in \Omega} \tau_{ij}\boldsymbol{w}_{jt} y_{ijt}+\Lambda_{u} \boldsymbol{\mu}_{u}\right),
    \end{aligned}
    \label{bttf_equation12}
\end{equation}
where $\boldsymbol{w}_{jt}=\boldsymbol{v}_{j}\circledast\boldsymbol{x}_{t}\in\mathbb{R}^{R}$. The conditional posterior distribution of $\boldsymbol{v}_{j}$ is defined in the same way.

Under the assumptions above, the full conditionals $p(\boldsymbol{\mu}_{u},\Lambda_{u}|-)$ and $p(\boldsymbol{\mu}_{v},\Lambda_{v}|-)$ are of the same Gaussian-Wishart form as $p(\boldsymbol{\mu}_{w},\Lambda_{w}|-)$ described in BTMF in Section \ref{sec:btmf_inference}. Similarly, the full conditional $p\left(A,\Sigma|-\right)$ is also of the same form as Eq.~\eqref{mniw_posterior} in BTMF.

For precision, we can simply adapt Eq.~\eqref{equ:tau} to sample $\tau_{ij}$ for each time series $(i,j)$. When assuming the noise is Gaussian isotropic, the posterior distribution of $\tau$ is $\text{Gamma}(\alpha^{*},\beta^{*})$ where
\begin{equation}
\begin{aligned}
\alpha^{*}&=\frac{1}{2}\sum\nolimits_{(i,j,t)\in\Omega}\mathbbm{1}(y_{ijt})+\alpha, \\
\beta^{*}&=\frac{1}{2}\sum\nolimits_{(i,j,t)\in\Omega}(y_{i,j,t}-\sum\nolimits_{r=1}^{R}u_{ir}v_{jr}x_{tr})^2+\beta. \\
\end{aligned}
\end{equation}

\subsection{Model Implementation}

We summarize the Gibbs sampling algorithm for missing data imputation of BTTF as Algorithm~\ref{bttf_algorithm}. For making rolling prediction with tensor data, we apply the same prediction mechanism of BTMF (see Algorithm~\ref{btmf_forecasting}) to adapt BTTF model.

\begin{algorithm}
 \caption{BTTF imputation}
 \label{bttf_algorithm}
\begin{algorithmic}[1]
 \renewcommand{\algorithmicrequire}{\textbf{Input:}}
 \renewcommand{\algorithmicensure}{\textbf{Output:}}
 \REQUIRE  data tensor $\mathcal{Y}\in\mathbb{R}^{M\times N\times T}$, $\Omega$ as the set of observed entries in $\mathcal{Y}$, $\mathcal{L}=\left\{h_1,h_2,\ldots,h_d\right\}$ as the time lag set, number of burn-in iterations $m_1$, and number of samples used in estimation $m_2$.
 Initialization of factor matrices $\left\{U,V,X\right\}$ and VAR coefficient matrix  $A\in\mathbb{R}^{Rd\times R}$.
 \ENSURE estimated tensor $\hat{\mathcal{Y}}\in\mathbb{R}^{M\times N\times T}$.
 \\ Initialize $\beta_0=1$, $\nu_0=R$, $\boldsymbol{\mu}_0=\boldsymbol{0}$ as a zero vector, $W_0=I_R$ ($S_0=I_{R}$ and $\Psi_0=I_{Rd}$) as an identity matrix, $M_0$ as all-zero mean matrix, and $\alpha,\beta=10^{-6}$.
  \FOR {$\text{iter.} = 1$ to $m_1+m_2$}
  \STATE Draw hyperparameters $\left\{\boldsymbol{\mu}_{u},\Lambda_{u},\boldsymbol{\mu}_{v},\Lambda_{v}\right\}$.
  \FOR {$i=1$ to $M$ (can be in parallel)}
  \STATE Draw $\boldsymbol{u}_{i}\sim\mathcal{N}(\boldsymbol{\mu}_{u}^{*},(\Lambda_{u}^{*})^{-1})$.
  \ENDFOR
  \FOR {$j=1$ to $N$ (can be in parallel)}
  \STATE Draw $\boldsymbol{v}_{j}\sim\mathcal{N}(\boldsymbol{\mu}_{v}^{*},(\Lambda_{v}^{*})^{-1})$.
  \ENDFOR
  \STATE Draw $\Sigma\sim\mathcal{IW}\left(S^{*},\nu^{*}\right)$ and $A\sim\mathcal{MN}\left(M^{*},\Psi^{*},\Sigma\right)$.
  \FOR {$t=1$ to $T$}
  \STATE Draw $\boldsymbol{x}_{t}\sim\mathcal{N}(\boldsymbol{\mu}_{t}^{*},\Sigma_{t}^{*})$.
  \ENDFOR
  \STATE Draw $\tau_{ij}$ or $\tau$ (isotropic noise) $\sim\text{Gamma}(\alpha^{*},\beta^{*})$.
  \IF {$\text{iter.} > m_1$}
  \STATE Compute $\tilde{\mathcal{Y}}=\sum_{r=1}^{R}\boldsymbol{u}_{r}\circ\boldsymbol{v}_{r}\circ\boldsymbol{x}_{r}$. Collect sample $\tilde{\mathcal{Y}}$.
  \ENDIF
  \ENDFOR
  \RETURN $\hat{\mathcal{Y}}$ as the average of the $m_2$ samples of $\tilde{\mathcal{Y}}$.
\end{algorithmic}
\end{algorithm}

\section{Experiments} \label{sec:experiments}

In this section we apply BTMF and BTTF on several real-world spatiotemporal data sets for both imputation and prediction tasks, and evaluate the effectiveness of these two models against recent state-of-the-art methods. We use the mean absolute percentage error (MAPE) and root mean square error (RMSE) as performance metrics:
\begin{equation*}
    \text{MAPE}=\frac{1}{n}\sum_{i=1}^{n}\frac{\vert y_i-\hat{y}_i\vert}{y_i}\times 100,\,
    \text{RMSE}=\sqrt{\frac{1}{n}\sum_{i=1}^{n}\left(y_i-\hat{y}_i\right)^2},
\end{equation*}
where $n$ is the total number of estimated values, and $y_i$ and $\hat{y}_i$ are the actual value and its estimation, respectively. For MCMC, we find the chains mix very fast on all data sets. The point estimates are obtained by averaging over $m_2=200$ Gibbs iterations. The Python code and adapted data sets for our experiments are publicly available at \url{https://github.com/xinychen/transdim}.

\begin{table*}[!ht]
\caption{Performance comparison (in MAPE/RMSE) for imputation tasks on data sets (G), (H), (S), and (L).}
\label{table1}
\centering
\begin{tabular}{l|l|rrrrrrrr}
\toprule
Data & Missing & {BTMF} & {BTRMF} & TRMF & {BPMF} & {BGCP} & {BATF} & HaLRTC \\
\midrule
\multirow{3}{*}{(G)} & 40\%, RM & 7.61/3.29 & \textbf{7.51}/\textbf{3.24} & 7.77/3.26 & 9.69/4.11 & 8.31/3.59 & 8.29/3.58 & 8.86/3.61 \\
& 60\%, RM & 8.20/3.51 & \textbf{7.95}/\textbf{3.42} & 8.48/3.52 & 10.19/4.30 & 8.42/3.64 & 8.43/3.64 & 9.82/3.96 \\
& 40\%, NM & 10.18/4.31 & 10.31/4.38 & 10.36/4.36 & 10.39/4.40 & 10.24/4.33 & \textbf{10.14}/\textbf{4.30} & 10.88/4.38 \\ \midrule
\multirow{3}{*}{(H)} & 40\%, RM & 24.0/\textbf{30.0} & 22.7/33.5 & 23.2/36.6 & 32.4/41.8 & 19.6/42.2 & 19.9/44.9 & \textbf{19.0}/31.8 \\
& 60\%, RM & 25.2/35.2 & 24.7/37.1 & 24.4/41.2 & 39.9/46.7 & 20.2/\textbf{34.9} & 20.5/38.8 & \textbf{20.1}/36.2 \\
& 40\%, NM & 25.9/46.0 & 27.1/51.7 & 26.0/\textbf{37.9} & 35.0/45.4 & \textbf{20.7}/45.6 & 21.2/40.5 & 21.5/53.1 \\ \midrule
\multirow{3}{*}{(S)} & 40\%, RM & 5.96/\textbf{3.75} & \textbf{5.95}/\textbf{3.75} & 6.17/3.80 & 6.79/4.18 & 7.42/4.49 & 7.40/4.47 & 6.76/3.83 \\
& 60\%, RM & 6.15/\textbf{3.83} & \textbf{6.13}/\textbf{3.83} & 6.48/3.92 & 7.38/4.49 & 7.47/4.52 & 7.45/4.50 & 7.90/4.34 \\
& 40\%, NM & 9.27/5.35 & 9.33/5.38 & \textbf{9.19}/5.30 & \textbf{9.19}/5.30 & 10.09/5.75 & 9.94/5.68 & 10.19/\textbf{5.27} \\ \midrule
\multirow{3}{*}{(L)} & 40\%, RM & 9.15/2.26 & 9.15/2.26 & 9.20/2.23 & 9.14/2.21 & 9.21/2.24 & 9.21/2.24 & \textbf{9.11}/\textbf{2.17} \\
& 60\%, RM & 9.32/2.30 & 9.32/2.30 & 9.36/2.27 & \textbf{9.28}/\textbf{2.24} & 9.34/2.27 & 9.33/2.27 & 9.51/2.27 \\
& 40\%, NM & \textbf{9.42}/2.32 & \textbf{9.42}/2.32 & 9.57/2.33 & 9.47/\textbf{2.30} & 9.54/2.33 & 9.55/2.33 & 9.84/2.35 \\
\bottomrule
\multicolumn{5}{l}{\scriptsize{Best results are highlighted in bold fonts.}}
\end{tabular}
\end{table*}

\begin{table*}[!ht]
\caption{Performance comparison (in MAPE/RMSE) for prediction tasks on data sets (G), (H), (S), and (L) with different time horizons.}
\label{table2}
\centering
\begin{tabular}{l|l|rrr|rrr|rrr}
\toprule
\multirow{2}{*}{Data} & \multirow{2}{*}{Missing} & \multicolumn{3}{c|}{BTMF} & \multicolumn{3}{c|}{BTRMF} & \multicolumn{3}{c}{TRMF} \\
\cmidrule(lr){3-5} \cmidrule(lr){6-8}
\cmidrule(lr){9-11}
& & $\delta=2$ & $\delta=4$ & $\delta=6$ & $\delta=2$ & $\delta=4$ & $\delta=6$ & $\delta=2$ & $\delta=4$ & $\delta=6$ \\
\midrule
\multirow{4}{*}{(G)} & Original & \textbf{11.01}/\textbf{4.44} & \textbf{11.22}/\textbf{4.54} & \textbf{11.59}/\textbf{4.68} & 11.37/4.58 & 11.81/4.75 & 12.34/4.94 & 11.60/4.55 & 11.96/4.61 & 12.28/4.90 \\
& 40\%, RM & \textbf{11.22}/\textbf{4.51} & \textbf{11.52}/\textbf{4.63} & \textbf{11.64}/\textbf{4.69} & 11.41/4.57 & 11.67/4.67 & 12.45/5.00 & 12.72/4.84 & 12.82/4.85 & 13.41/5.14 \\
& 60\%, RM & \textbf{11.44}/\textbf{4.58} & \textbf{11.66}/\textbf{4.66} & \textbf{11.93}/\textbf{4.78} & 11.68/4.67 & 12.23/4.82 & 12.64/5.01 & 13.85/5.19 & 13.94/5.19 & 14.30/5.33 \\
& 40\%, NM & \textbf{11.35}/\textbf{4.60} & \textbf{11.64}/\textbf{4.65} & \textbf{11.80}/\textbf{4.75} & 11.99/4.77 & 12.26/4.93 & 12.96/5.19 & 12.80/4.88 & 12.84/4.88 & 13.32/5.11 \\
\midrule
\multirow{4}{*}{(H)} & Original & 25.4/34.7 & 27.3/36.0 & 27.7/37.5 & 25.7/34.1 & 27.6/36.2 & 31.0/38.0 & \textbf{22.5}/\textbf{30.6} & \textbf{24.1}/\textbf{32.6} & \textbf{25.4}/\textbf{33.9} \\
& 40\%, RM & 27.0/38.1 & \textbf{26.6}/40.1 & 27.4/42.7 & 30.0/40.0 & 29.8/41.9 & 30.2/42.5 & \textbf{23.8}/\textbf{35.8} & 27.3/\textbf{38.4} & \textbf{27.3}/\textbf{40.4} \\
& 60\%, RM & 26.8/\textbf{41.1} & 27.2/\textbf{42.3} & 28.5/\textbf{43.6} & 30.4/43.2 & 29.6/43.9 & 31.7/44.5 & \textbf{25.8}/41.9 & \textbf{25.1}/44.2 & \textbf{26.8}/46.5 \\
& 40\%, NM & 27.5/48.6 & 29.2/45.3 & 30.7/47.7 & 29.8/42.5 & 31.7/49.4 & 39.7/50.5 & \textbf{25.6}/\textbf{38.6} & \textbf{28.0}/\textbf{40.4} & \textbf{27.7}/\textbf{42.6} \\
\midrule
\multirow{4}{*}{(S)} & Original & \textbf{10.39}/5.63 & 10.82/5.83 & 11.18/6.02 & 11.44/6.21 & 12.30/6.60 & 12.97/7.00 & 10.45/\textbf{5.58} & \textbf{10.65}/\textbf{5.70} & \textbf{11.15}/\textbf{5.92} \\
& 40\%, RM & \textbf{10.52}/\textbf{5.69} & \textbf{10.96}/\textbf{5.89} & \textbf{11.37}/\textbf{6.09} & 11.58/6.25 & 12.34/6.62 & 12.79/6.93 & 11.37/5.85 & 11.53/5.96 & 11.98/6.15 \\
& 60\%, RM & 11.60/6.30 & \textbf{11.21}/\textbf{6.10} & \textbf{12.80}/7.03 & \textbf{11.58}/6.24 & 12.36/6.62 & 12.99/6.99 & 12.38/\textbf{6.19} & 12.67/6.37 & 12.88/\textbf{6.42} \\
& 40\%, NM & \textbf{10.93}/\textbf{5.96} & \textbf{11.17}/6.03 & \textbf{11.55}/6.24 & 11.66/6.35 & 12.54/6.73 & 13.38/7.29 & 11.63/5.98 & 11.68/\textbf{6.02} & 12.00/\textbf{6.17} \\
\midrule
\multirow{4}{*}{(L)} & Original & \textbf{11.21}/\textbf{2.67} & \textbf{11.36}/\textbf{2.70} & \textbf{12.01}/\textbf{2.82} & 12.69/3.17 & 14.82/3.89 & 18.10/4.09 & 11.35/2.68 & 11.64/2.79 & 12.80/3.05 \\
& 40\%, RM & \textbf{11.41}/\textbf{2.70} & \textbf{11.54}/\textbf{2.73} & \textbf{12.15}/\textbf{2.84} & 12.25/3.03 & 12.35/2.98 & 15.68/4.03 & 11.49/2.71 & 11.70/2.76 & 12.60/2.95 \\
& 60\%, RM & \textbf{11.53}/\textbf{2.72} & \textbf{11.69}/\textbf{2.75} & \textbf{12.28}/\textbf{2.86} & 12.42/3.01 & 13.21/3.25 & 14.87/3.70 & 11.95/2.79 & 12.02/2.80 & 12.68/2.95 \\
& 40\%, NM & \textbf{11.37}/\textbf{2.71} & \textbf{11.55}/\textbf{2.74} & \textbf{12.11}/\textbf{2.85} & 13.80/3.47 & 14.44/3.65 & 17.54/4.40 & 11.62/2.72 & 12.78/3.14 & 12.30/2.87 \\
\bottomrule
\multicolumn{4}{l}{\scriptsize{Best results are highlighted in bold fonts.}}
\end{tabular}
\end{table*}

\subsection{BTMF}

\noindent\textbf{Data set (G): Guangzhou urban traffic speed\footnote{\url{https://doi.org/10.5281/zenodo.1205229}}}. This data set registered traffic speed data from 214 road segments over two months (61 days from August 1 to September 30, 2016) with a 10-minute resolution (144 time intervals per day) in Guangzhou, China. We organize the raw data set into a time series matrix of $214\times 8784$ and there are 1.29\% missing values.

\noindent\textbf{Data set (H): Hangzhou metro passenger flow\footnote{\url{https://tianchi.aliyun.com/competition/entrance/231708/information}}}. This data set collected incoming passenger flow from 80 metro stations over 25 days (from January 1 to January 25, 2019) with a 10-minute resolution in Hangzhou, China. We discard the interval 0:00 a.m. -- 6:00 a.m. with no services (i.e., only consider the remaining 108 time intervals) and re-organize the raw data set into a time series matrix of $80\times 2700$. { The flow data is highly heterogeneous, with min=0, max=3334, and mean=135.}

\noindent\textbf{Data set (S): Seattle freeway traffic speed\footnote{\url{https://github.com/zhiyongc/Seattle-Loop-Data}}}. This data set collected freeway traffic speed from 323 loop detectors with a 5-minute resolution over the whole year of 2015 in Seattle, USA. We choose the data from January 1 to January 28 (i.e., 4 weeks) as our experiment data, and organize the data set into a time series matrix matrix of $323\times 8064$.

{

\noindent\textbf{Data set (L): London movement speed}. This data set is created by Uber movement project\footnote{\url{https://movement.uber.com/}}, which includes the average speed on a given road segment for each hour of each day in April 2019. In this data set, there are about 220,000 road segments. Note that this data set only includes road segments with at least 5 unique trips in that hour. There are up to 73.09\% missing values in total and most missing values are produced during night. We choose the subset of this raw data set into a time series matrix of $35912\times720$ in which each time series has at least 70\% observations.
}

\begin{figure*}[ht!]
\centering
\subfigure[Movement speed of road segment \#1.]{
    \centering
    \includegraphics[scale=0.46]{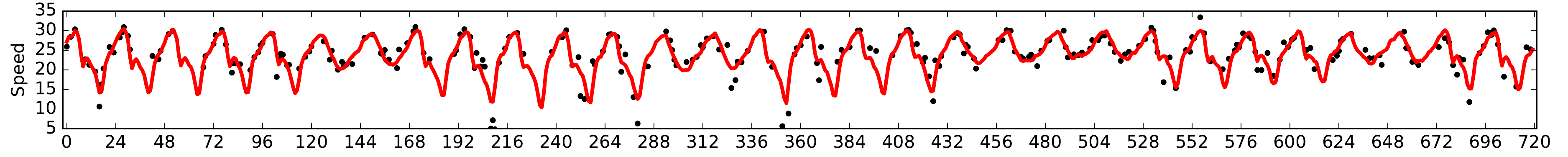}
}
\subfigure[Movement speed of road segment \#2.]{
    \centering
    \includegraphics[scale=0.46]{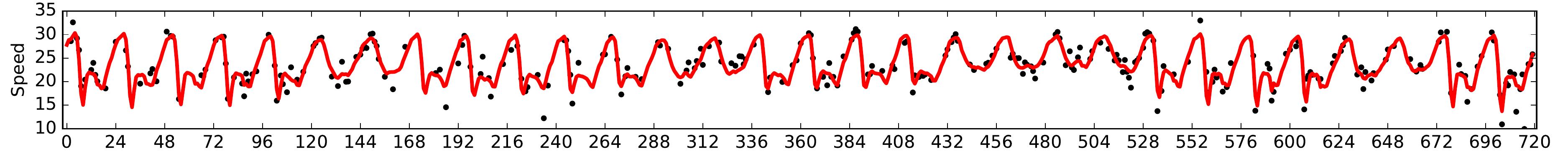}
}
\subfigure[Movement speed of road segment \#3.]{
    \centering
    \includegraphics[scale=0.46]{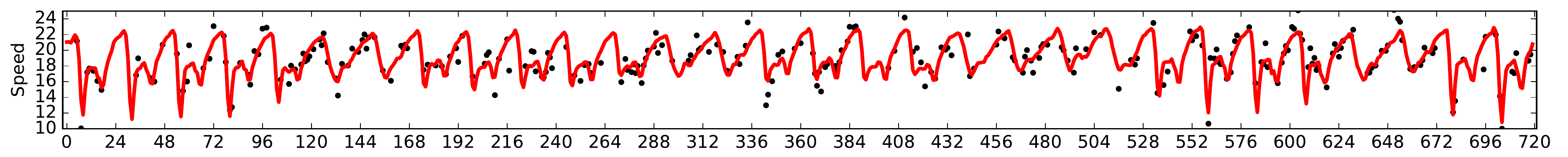}
}
\caption{Imputed speed (mph) of BTMF on data set (L) with 60\% RM. Black dots show the partially observed (training) data, while red curves show the imputed speed values by BTMF.}
\label{imputation_london}
\end{figure*}

\noindent\textbf{Baselines}. We choose 1) TRMF \cite{yu2016temporal} and 2) its fully Bayesian counterpart---BTRMF---as main baseline models. Note that BTRMF can be considered a special case of BTMF with the independent factor constraint in Eq.~\eqref{equ:AR} instead of a VAR in Eq.~\eqref{equ:VAR} (i.e., restricting $A_k$ to be diagonal). Similar to BTMF, we use a spatially-varying precision parameter $\tau_i$ to characterize the noise level of time series $\boldsymbol{y}_{i}$ in BTRMF with the exception of (H). { Given the great heterogeneity in Hangzhou data, BTMF may give a large variance $\tau_i^{-1}$ on a particular time series. To avoid this issue, we assume the noises are both Gaussian isotropic in BTMF and BTRMF on (H).} We also consider a family of tensor-based models for missing data imputation, including: 3) Bayesian Gaussian CP decomposition (BGCP) \cite{chen2019abayesian}, which is a high-order extension of BPMF \cite{salakhutdinov2008bayesian}; 4) Bayesian Augmented Tensor Factorization (BATF) \cite{chen2019missing}; 5) HaLRTC: High-accuracy Low-Rank Tensor Completion \cite{liu2013tensor}. These models are chosen because matrix time series data collected from multiple days can be re-organized as a third-order (location$\times$day$\times$time of day) tensor, and in this case tensor factorization can effectively learn the global patterns provided by the additional ``day'' dimension. In fact, these tensor models have shown superior performance in various imputation tasks (e.g., traffic data and images). For prediction, we compare BTMF against TRMF and BTRMF. In doing so, we adapt TRMF/BTRMF to an implementation which is similar to the scheme of BTMF forecasting.

\noindent\textbf{Experiment setup.}
We assess the performance of these models under two common missing data scenarios---random missing (\textbf{RM}) and non-random missing (\textbf{NM}). For RM, we simply remove a certain amount of observed entries in the matrix randomly and use these entries as ground truth to evaluate MAPE and RMSE. The percentages of missing values are set as 40\%/60\% for all these data sets. For NM, we apply a fiber/block missing experiment by randomly choosing certain location$\times$day combinations and removing all the observations in each combination. The percentages of missing values are set as 40\%. Again, the removed but actually observed entries are used for evaluation. The NM scenario corresponds to the cases where sensors have a certain probability to fail on each day. For tensor-based baseline models (BGCP, BATF, and HaLRTC), we re-organize the matrix into a third-order (location$\times$day$\times$time slot) tensor as input. For matrix based models, we use the original time series matrix (location$\times$time series) as input. For BTMF, BTRMF, and TRMF, we use a small time lag set $\mathcal{L}=\left\{1,2,T_0\right\}$ for all data sets, where $T_0$ denotes the number of time slots per day. For TRMF, we perform grid search from \{50, 5, 0.5, 0.05\} for each regularization parameter as in \cite{yu2016temporal}. We set the learning rate in HaLRTC as $\rho=10^{-5}$. We set the minimum low rank as $R=10$ for all the factorization models.

For prediction tasks, we apply some multi-step rolling prediction experiments as described in Section~\ref{sec:btmf_implementation}. We evaluate these models by making rolling predictions over last seven days for all these data sets. Total number of predicted time slots are $7\times 144$, $7\times 108$, $7\times 288$, and $7\times 24$ for
data set (G), (H), (S), and (L), respectively. For data set (G), (H), (S), and (L), we set the time horizon $\delta$ of each rolling prediction as $\{2,4,6\}$. To guarantee model performance on multi-step prediction tasks, we set time lags as $\mathcal{L}=\{1,2,3,T_0,T_0+1,T_0+2,7T_0,7T_0+1,7T_0+2\}$. Note that BTMF, BTRMF, and TRMF does not impute missing values for these prediction tasks. The low rank of these models is set as $R=10$.

\begin{figure*}[!ht]
\centering
\subfigure[Metro station \#1.]{
    \centering
    \includegraphics[scale=0.5]{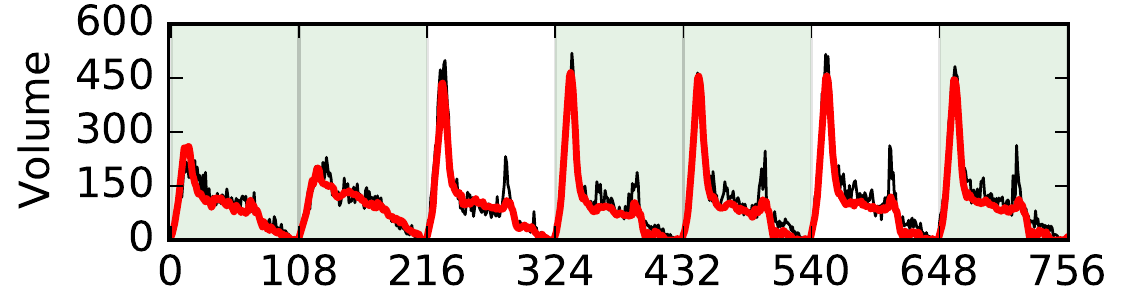}
}
\subfigure[Metro station \#2.]{
    \centering
    \includegraphics[scale=0.5]{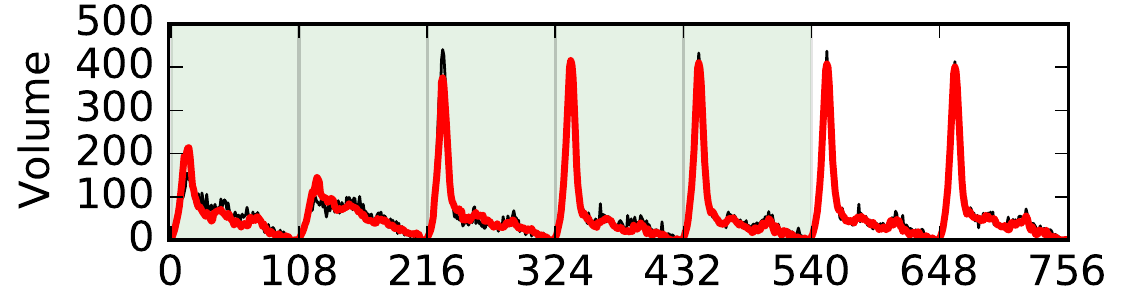}
}
\subfigure[Metro station \#3.]{
    \centering
    \includegraphics[scale=0.5]{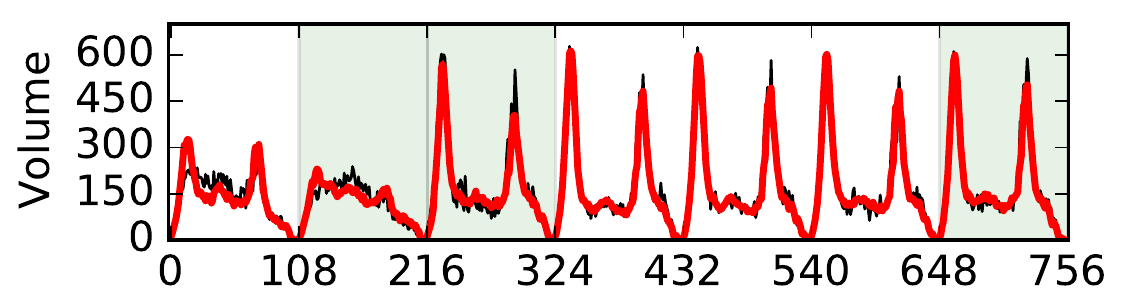}
}
\subfigure[Metro station \#31.]{
    \centering
    \includegraphics[scale=0.5]{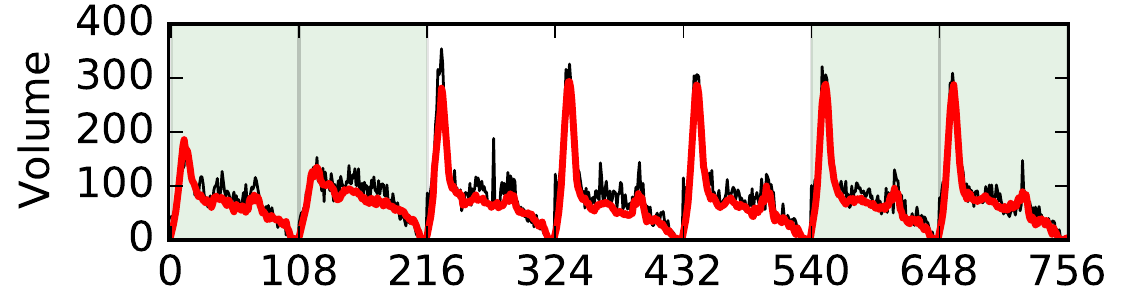}
}
\subfigure[Metro station \#32.]{
    \centering
    \includegraphics[scale=0.5]{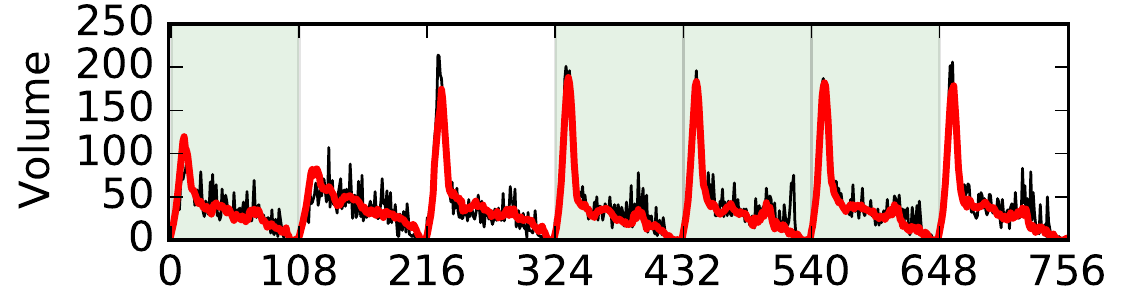}
}
\subfigure[Metro station \#33.]{
    \centering
    \includegraphics[scale=0.5]{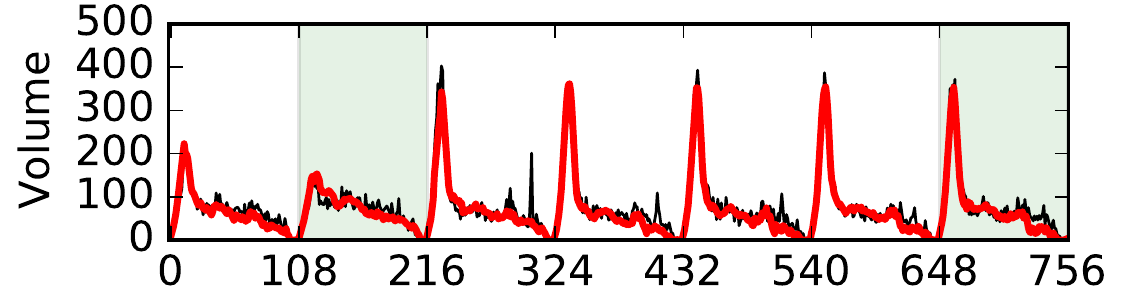}
}
\subfigure[Metro station \#61.]{
    \centering
    \includegraphics[scale=0.5]{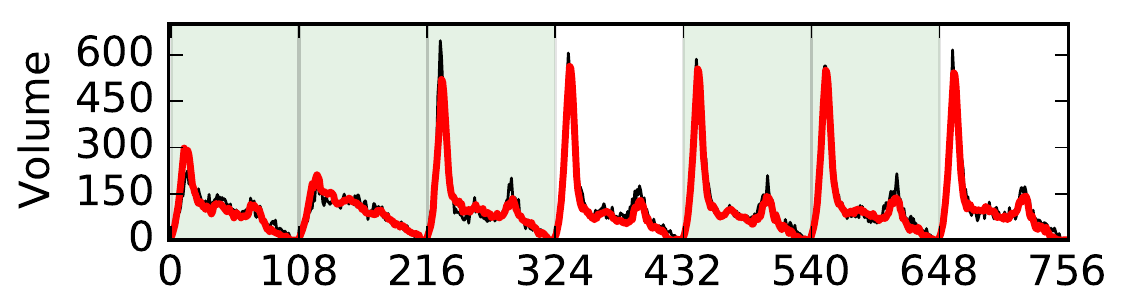}
}
\subfigure[Metro station \#62.]{
    \centering
    \includegraphics[scale=0.5]{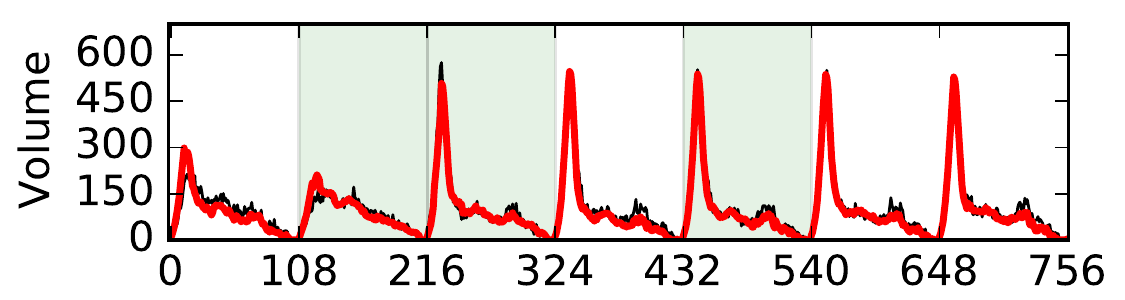}
}
\subfigure[Metro station \#63.]{
    \centering
    \includegraphics[scale=0.5]{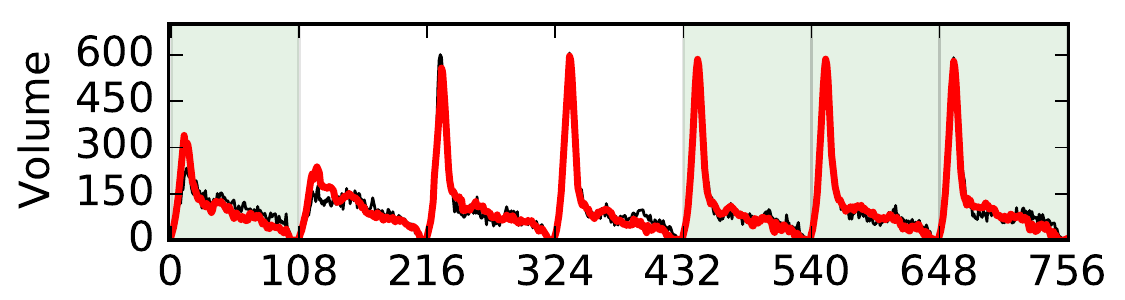}
}
\caption{Predicted metro passenger flow (red curves) of BTMF (time horizon: $\delta=6$) at 40\% NM missing scenario vs. actual observations (black curves) for data set (H). In these panels, white rectangles represent non-random missing (i.e., volume observations are lost in a whole day).}
\label{prediction_Hangzhou}
\end{figure*}

\begin{figure*}[ht!]
\centering
\subfigure[Road segment \#1.]{
\centering
\includegraphics[scale=0.5]{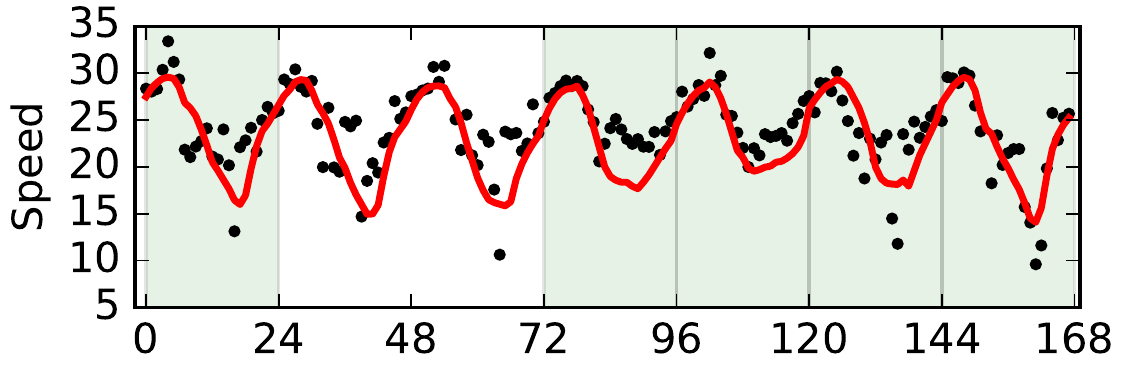}
}
\subfigure[Road segment \#2.]{
\centering
\includegraphics[scale=0.5]{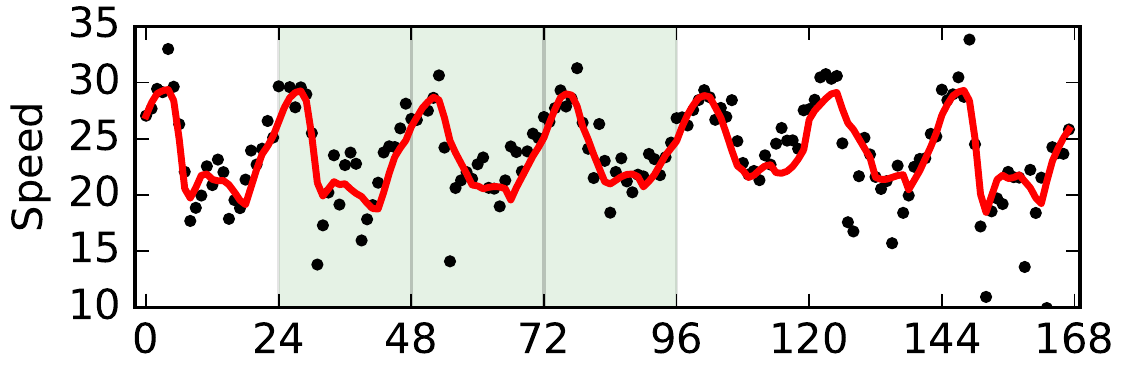}
}
\subfigure[Road segment \#3.]{
\centering
\includegraphics[scale=0.5]{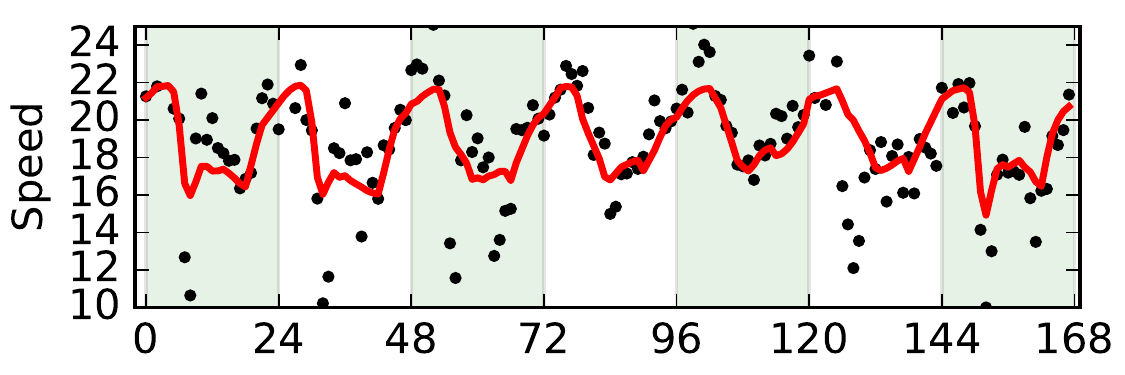}
}
\subfigure[Road segment \#4.]{
\centering
\includegraphics[scale=0.5]{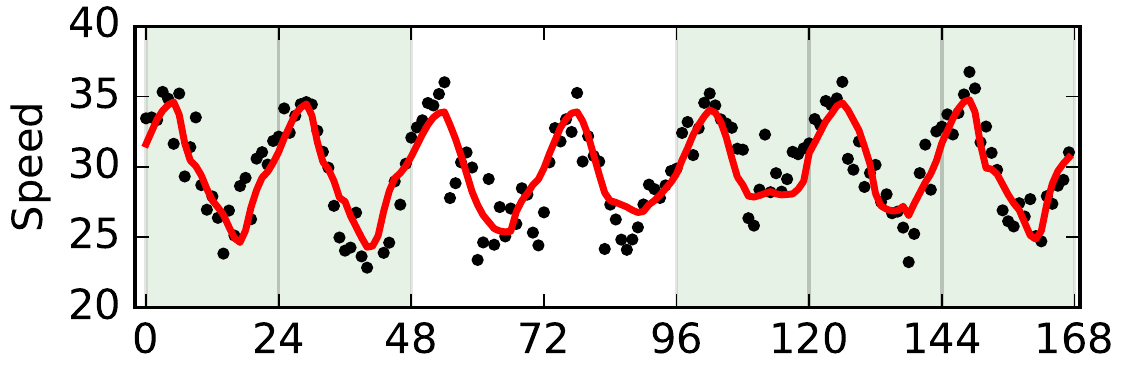}
}
\subfigure[Road segment \#5.]{
\centering
\includegraphics[scale=0.5]{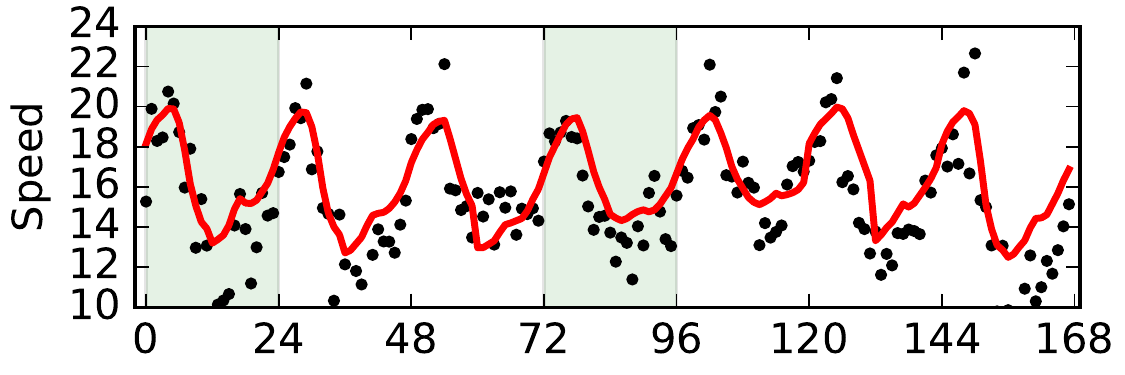}
}
\subfigure[Road segment \#6.]{
\centering
\includegraphics[scale=0.5]{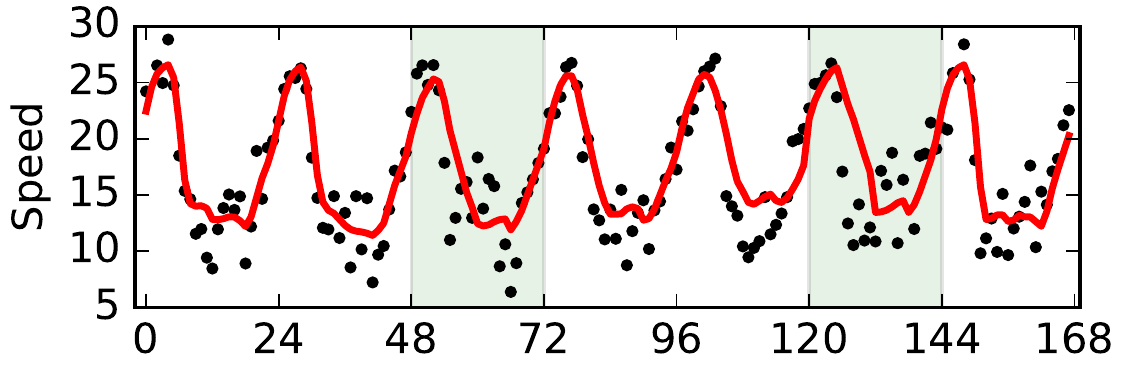}
}
\caption{Predicted movement speed (red curves) of BTMF (time horizon: $\delta=6$) at 40\% NM missing scenario vs. actual observations (black dots) for data set (L). In these panels, white rectangles represent non-random missing (i.e., speed observations are lost in a whole day).}
\label{prediction_London}
\end{figure*}

\noindent\textbf{Results and analysis}. Table~\ref{table1} shows the imputation performance of BTMF and other baseline models for data sets (G), (H), (S), and (L). The results in all experiments are given by ``MAPE/RMSE''. Of the reported MAPE/RMSE, non-random missing values seem to be more difficult to reconstruct with all these imputation models than random missing values. As can be seen, the proposed BTMF and the adapted BTRMF clearly outperform TRMF in most cases. The results also reveal that Bayesian treatment over temporal matrix factorization is more superior than manually controlling the regularizers. Essentially, among all matrix models, the imputation performance of BPMF is inferior to the temporal matrix factorization models---BTMF, BTRMF, and TRMF---especially in RM scenarios because the local temporal consistency is ignored in BPMF. Tensor models like BGCP, BATF, and HaLRTC also achieve competitive results due to the global temporal consistency introduced by the ``day" dimension. Our results suggest that BTMF (or BTRMF) inherits the advantages of both matrix models (e.g., TRMF and BPMF) and tensor models (e.g., BGCP and BATF) even with a very simple time lag set: It not only provides a flexible and automatic inference technique for model parameter estimation, but also offers superior imputation performance by integrating temporal dynamics into matrix factorization. As shown in Fig.~\ref{imputation_london}, our proposed BTMF can get the true signals of movement speed values and accurate imputations on the sparsely sensed data.

We next conduct the experiment for making multi-step rolling prediction on the four data sets. Table~\ref{rolling_prediction} shows the prediction performance of BTMF and other baseline models. As we can see, BTMF outperforms TRMF and BTRMF in most cases. Comparing BTRMF and TRMF, we find that the well tuned TRMF actually performs better than BTRMF in most cases. A possible reason is that we have performed extensive tuning to find the right regularization parameters for TRMF. Thus, the comparision may not be fair since we only run BTRMF once with non-informative priors.

The most interesting analysis is to compare BTMF and BTRMF. Both models are Bayesian with identical priors. The only difference is that we restrict $A_k$ to be diagonal in BTRMF (independent factor assumption in Eq.~\eqref{equ:AR}) while the VAR dynamics in Eq.~\eqref{equ:VAR} is integrated in BTMF. The results in Table~\ref{rolling_prediction} show that BTMF consistently outperform BTRMF. This further verifies the superiority of introducing the VAR prior on the temporal latent factor, which has better performance in characterizing the covariance structures and causal relationships than the simple AR mechanism. We next show some visualizations on the prediction performance of BTMF. As shown in Fig.~\ref{prediction_Hangzhou}, our proposed BTMF achieves very accurate prediction results on the Hangzhou metro passenger flow time series data, and the accuracy can be guaranteed even a considerable amount (i.e., 40\%) of the input sequence is missing. Some examples of forecasting results under severe missingness are provided in Fig.~\ref{prediction_Hangzhou}(c) and (f). Fig.~\ref{prediction_London} shows an example on the prediction results from BTMF on the London movement data. As we can see, BTMF not only learns the right temporal dynamics from partially observed data, but also can mitigate the impacts of large noise.

\begin{table*}[!ht]
\caption{Performance comparison (in MAPE/RMSE) for imputation tasks on data sets (N) and (P).}
\label{BTTF_imputation}
\centering
\begin{tabular}{l|l|rrrrrrr}
\toprule
Data & Missing & BTTF & BTRTF & BPTF &  BGCP & BATF & HaLRTC \\
\midrule
\multirow{3}{*}{(N)} & 40\%, RM & \textbf{47.8}/\textbf{4.85} & 48.5/4.86  & 48.0/4.91 & 48.9/4.82 & 49.3/4.90 & 51.0/6.84 \\
& 60\%, RM & 48.9/5.08 & \textbf{48.5}/5.03 & 49.3/\textbf{5.02} & 48.7/5.11 & 50.1/5.12 & 52.1/8.13 \\
& 40\%, NM & 48.7/5.05 & 48.8/5.03 & \textbf{48.5}/5.35 & 52.9/4.87 & 54.7/\textbf{4.86} & 51.5/7.03 \\
\midrule
\multirow{3}{*}{(P)} & 40\%, RM & 1.48/0.49 & 1.47/0.49 & 1.65/0.69 & 1.48/0.49 & 1.43/0.48 & \textbf{0.67}/\textbf{0.24} \\
& 60\%, RM & 1.49/0.50 & 1.47/0.49 & 1.68/0.69 & 1.48/0.50 & 1.43/0.48 & \textbf{0.98}/\textbf{0.35} \\
& 40\%, NM & 1.47/0.49 & 1.47/0.49 & 1.65/0.70 & 1.46/0.49 & 1.45/0.48 & \textbf{0.72}/\textbf{0.25} \\
\bottomrule
\multicolumn{5}{l}{\scriptsize{Best results are highlighted in bold fonts.}}
\end{tabular}
\end{table*}

\begin{table*}[!ht]
\caption{Performance comparison (in MAPE/RMSE) for prediction tasks on data sets (N) and (P) with different time horizons.}
\label{bttf_forecasting_result}
\centering
\begin{tabular}{l|l|rrr|rrr}
\toprule
\multirow{2}{*}{Data} & \multirow{2}{*}{Missing} & \multicolumn{3}{c|}{BTTF} & \multicolumn{3}{c}{BTRTF} \\
\cmidrule(lr){3-5} \cmidrule(lr){6-8}
& & $\delta=2$ & $\delta=4$ & $\delta=6$ & $\delta=2$ & $\delta=4$ & $\delta=6$ \\
\midrule
\multirow{4}{*}{(N)} & Original & 53.9/\textbf{5.10} & 54.7/\textbf{5.18} & 56.2/\textbf{5.25}  & \textbf{50.7}/5.47 & \textbf{50.9}/5.72 & \textbf{50.3}/5.94 \\
& 40\%, RM & \textbf{48.2}/\textbf{5.95} & 53.7/\textbf{5.33} & 56.1/\textbf{5.40} & 48.5/6.47 & \textbf{50.7}/5.75 & \textbf{50.9}/5.92 \\
& 60\%, RM & \textbf{48.9}/\textbf{6.73} & 53.0/\textbf{5.43} & 55.3/\textbf{5.46} & 49.3/7.28 & \textbf{50.6}/5.95 & \textbf{51.0}/6.16 \\
& 40\%, NM & \textbf{48.1}/\textbf{6.15} & 54.9/\textbf{5.32} & 54.8/\textbf{5.36} & 48.6/6.30 & \textbf{50.4}/5.88 & \textbf{50.1}/6.12 \\
\midrule
\multirow{4}{*}{(P)} & Original & \textbf{2.85}/\textbf{0.92} & \textbf{2.58}/\textbf{0.84} & \textbf{2.81}/\textbf{0.91} & 16.33/5.59 & 21.94/7.52 & 17.25/5.59 \\
& 40\%, RM & \textbf{2.31}/\textbf{0.76} & \textbf{2.50}/\textbf{0.81} & \textbf{2.58}/\textbf{0.84} & 10.07/3.30 & 14.66/4.71 & 14.57/4.71 \\
& 60\%, RM & \textbf{2.35}/\textbf{0.76} & \textbf{2.43}/\textbf{0.79} & \textbf{3.22}/\textbf{1.03} & 17.57/5.57 & 13.33/4.16 & 9.07/3.14 \\
& 40\%, NM & \textbf{2.32}/\textbf{0.75} & \textbf{2.41}/\textbf{0.79} & \textbf{2.45}/\textbf{0.80} & 28.51/9.45 & 10.32/3.34 & 29.62/9.49 \\
\bottomrule
\multicolumn{4}{l}{\scriptsize{Best results are highlighted in bold fonts.}}
\end{tabular}
\end{table*}

\subsection{BTTF}

\textbf{Data set (N): NYC taxi\footnote{\url{https://www1.nyc.gov/site/tlc/about/tlc-trip-record-data.page}}}. This data set registered trip information (pick-up/drop-off locations and start time) for different types of taxi trips. For the experiment, we choose the trips collected during May and June, 2018 (61 days) and organize the raw data into a third-order (pick-up zone$\times$drop-off zone$\times$time slot) tensor. We define in total 30 pick-up/drop-off zones and the temporal resolution for aggregating trips is selected as 1h. The size of this spatiotemporal tensor is $30\times 30\times 1464$.

{
\noindent\textbf{Data set (P): Pacific surface temperature\footnote{\url{http://iridl.ldeo.columbia.edu/SOURCES/.CAC/}}}. This data set collected monthly sea surface temperature on the Pacific over 396 consecutive months from January 1970 to December 2002. The spatial locations are expressed as grids of 2-by-2 degrees. The grid amount is $30\times 84$, and as a result, the temperature tensor is of size $30\times 84\times 396$.
}

\noindent\textbf{Baselines}. For the imputation tasks, we select some baseline models, including 1) fully Bayesian model of Temporal Regularized Tensor Factorization (BTRTF) with independent temporal factors (Eq.~\eqref{equ:AR}), as a natural higher-order extension of BTRMF. Other baseline imputation models include 2) BPTF \cite{xiong2010temporalcf}, 3) BGCP, 4) BATF, and 5) HaLRTC, which also have been used above. For the prediction tasks, we mainly compare the proposed BTTF with BTRTF. Note that TRTF---the tensor extension of TRMF---is not included as a baseline. This is because the model has five regularization parameters to tune, which is very challenging.

\begin{figure*}[!ht]
\centering
\subfigure[From zone 17 to zone 13.]{
    \centering
    \includegraphics[scale=0.53]{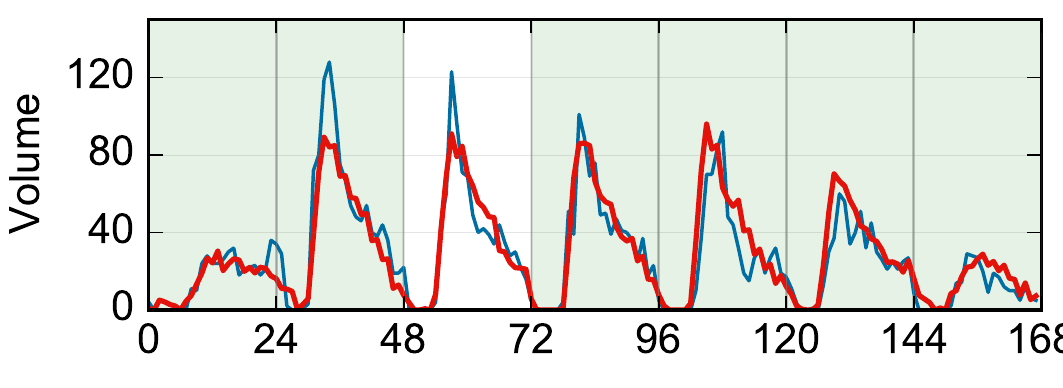}
}
\subfigure[From zone 17 to zone 21.]{
    \centering
    \includegraphics[scale=0.53]{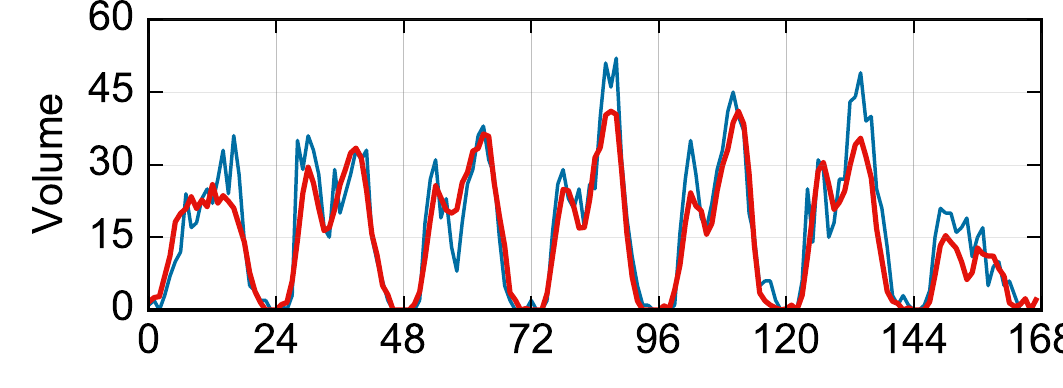}
}
\subfigure[From zone 17 to zone 21.]{
    \centering
    \includegraphics[scale=0.53]{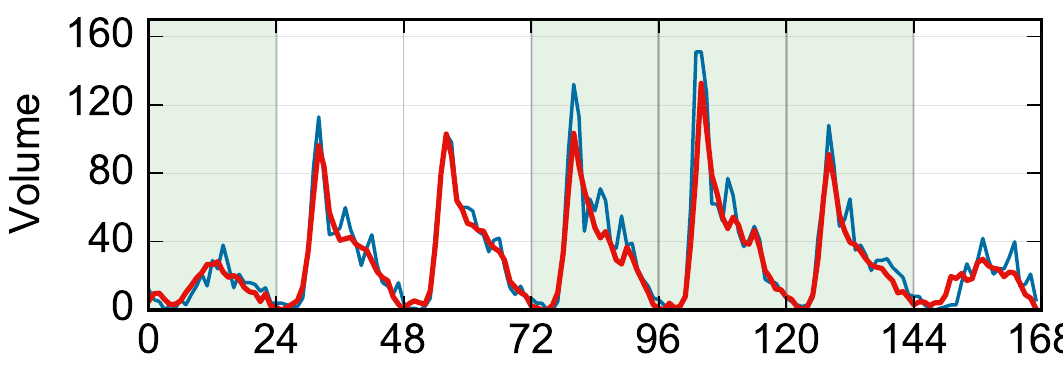}
}
\subfigure[From zone 17 to zone 27.]{
    \centering
    \includegraphics[scale=0.53]{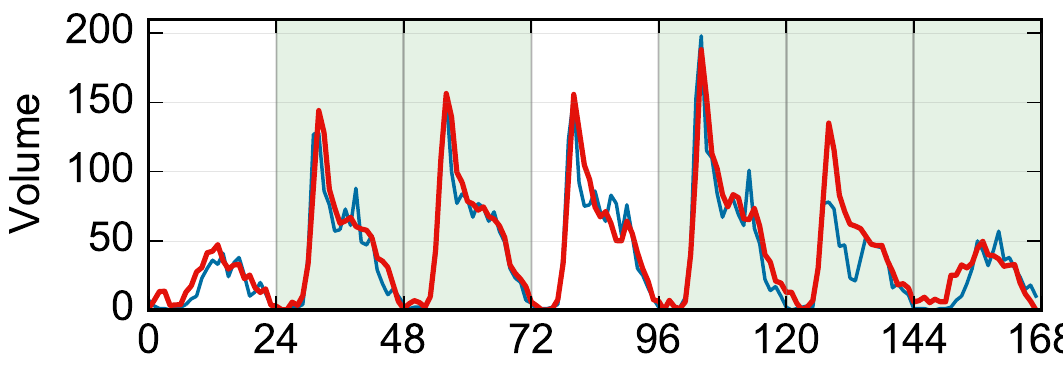}
}
\subfigure[From zone 26 to zone 21.]{
    \centering
    \includegraphics[scale=0.53]{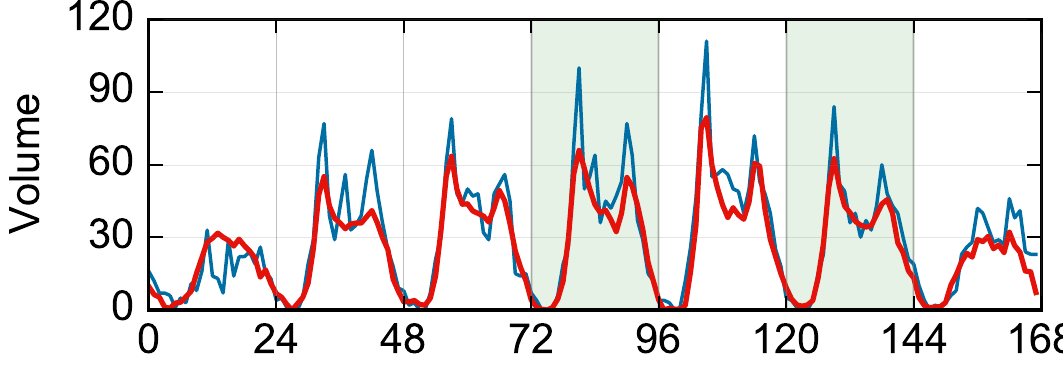}
}
\subfigure[From zone 27 to zone 27.]{
    \centering
    \includegraphics[scale=0.53]{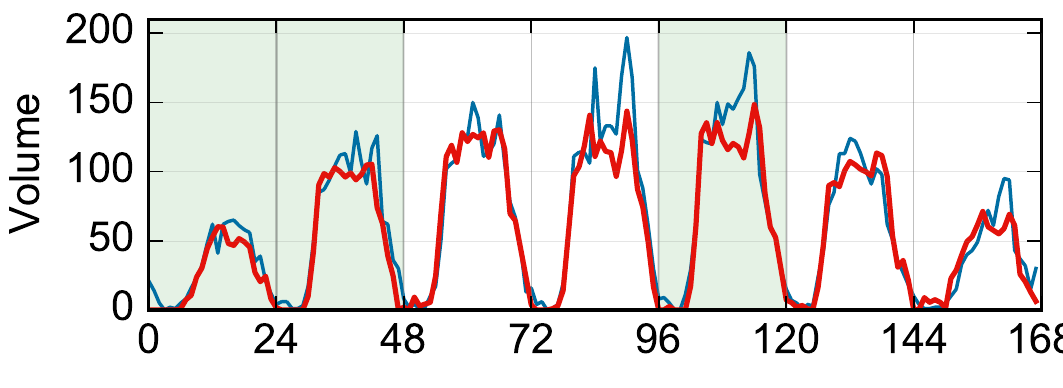}
}
\caption{Examples of six pick-up/drop-off pairs. We show the predicted time series using BTTF (time horizon: $\delta=2$) under 40\% NM (red curves) and the actual observations (blue curves). In these panels, white rectangles represent non-random missing.}
\label{nyc_prediction_example_time_series}
\end{figure*}

\begin{figure*}[!ht]
\centering
\subfigure[Actual volume.]{
    \centering
    \includegraphics[scale=0.57]{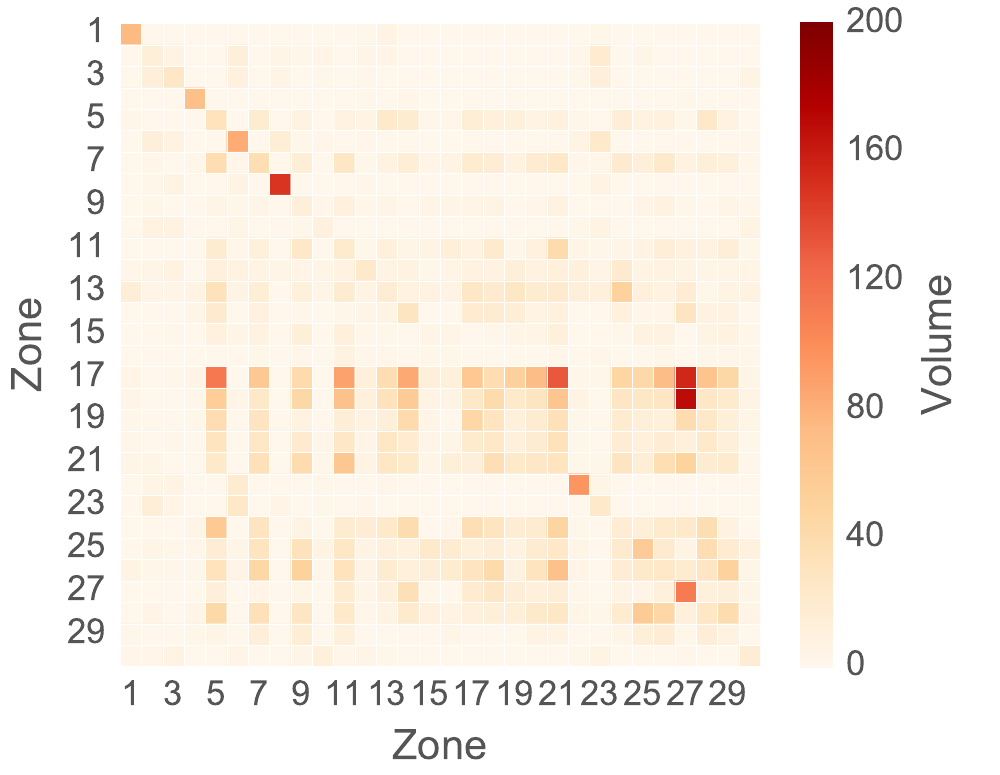}
}
\subfigure[Predicted volume with original data.]{
    \centering
    \includegraphics[scale=0.57]{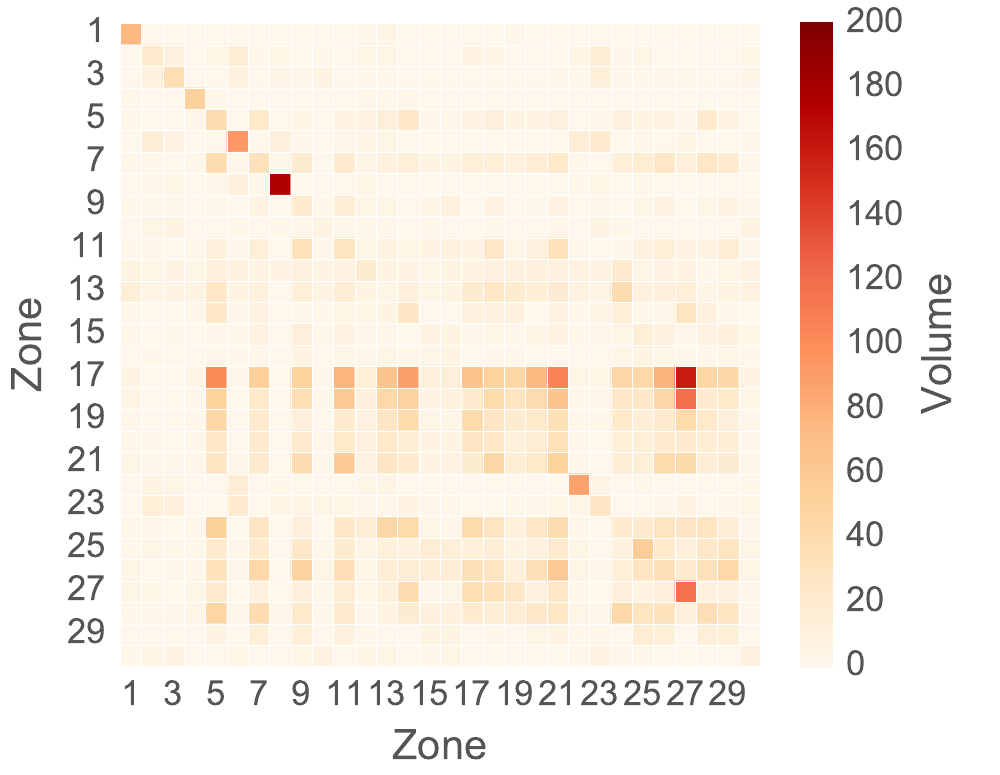}
}
\subfigure[Predicted volume with 40\% NM data.]{
    \centering
    \includegraphics[scale=0.57]{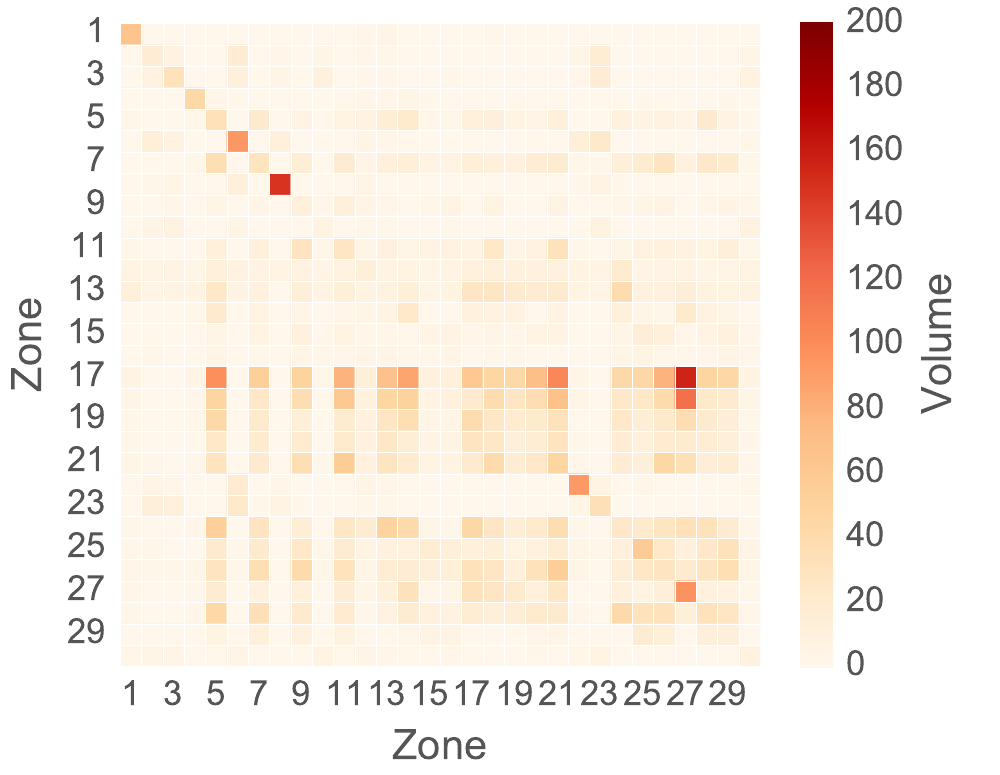}
}
\subfigure[Actual volume.]{
    \centering
    \includegraphics[scale=0.57]{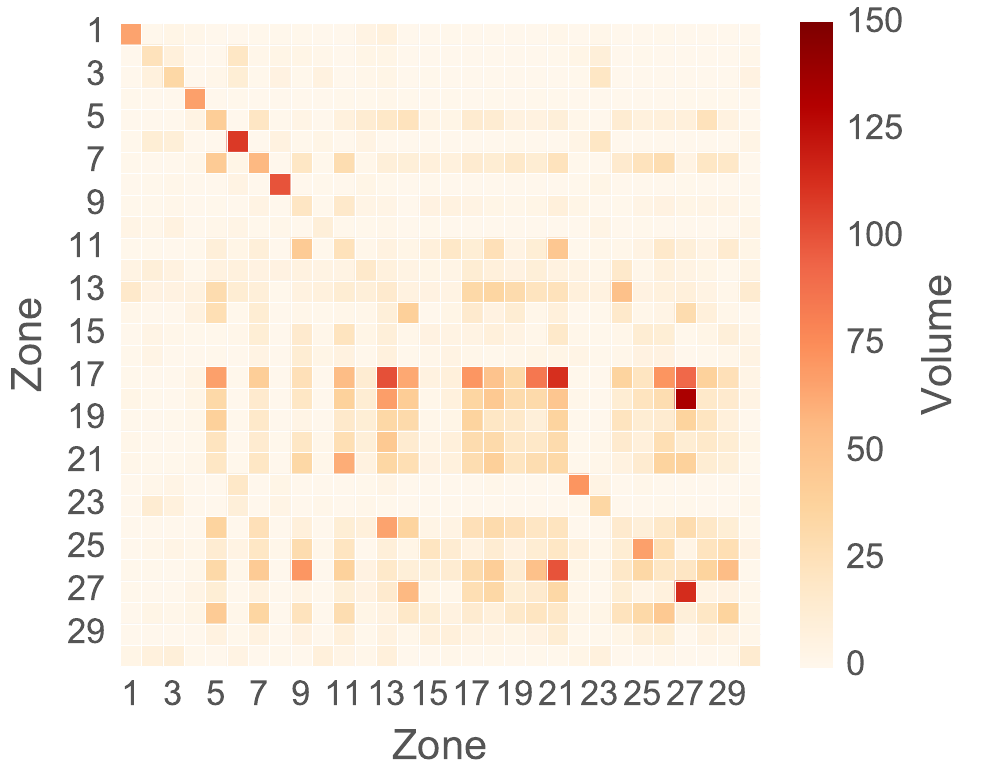}
}
\subfigure[Predicted volume with original data.]{
    \centering
    \includegraphics[scale=0.57]{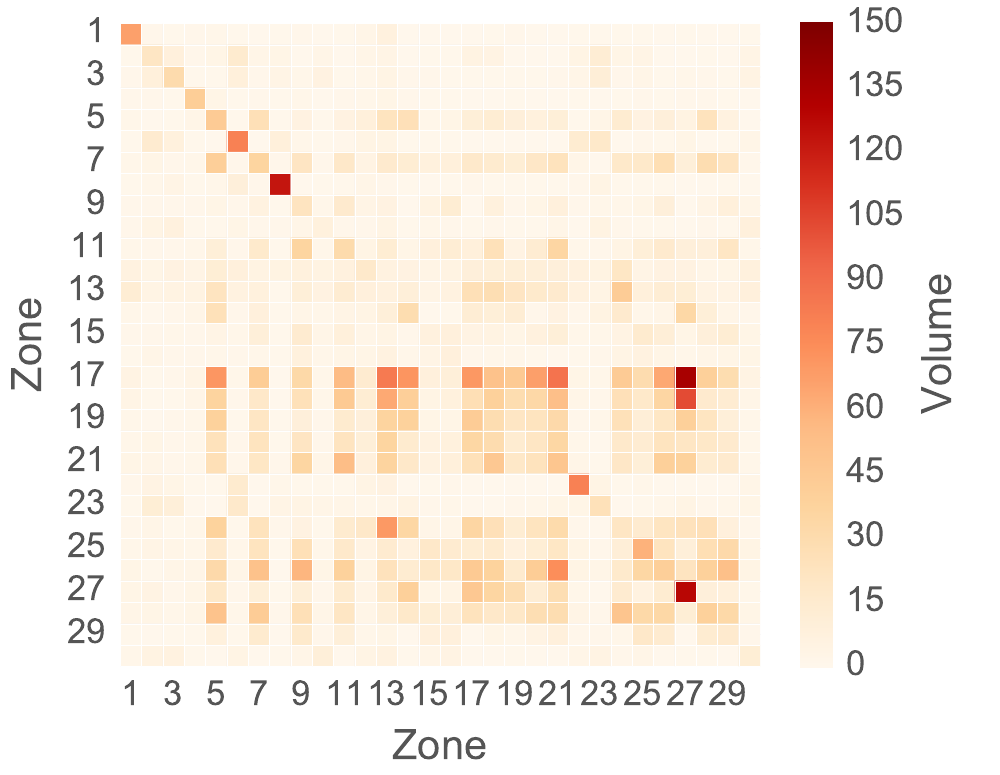}
}
\subfigure[Predicted volume with 40\% NM data.]{
    \centering
    \includegraphics[scale=0.57]{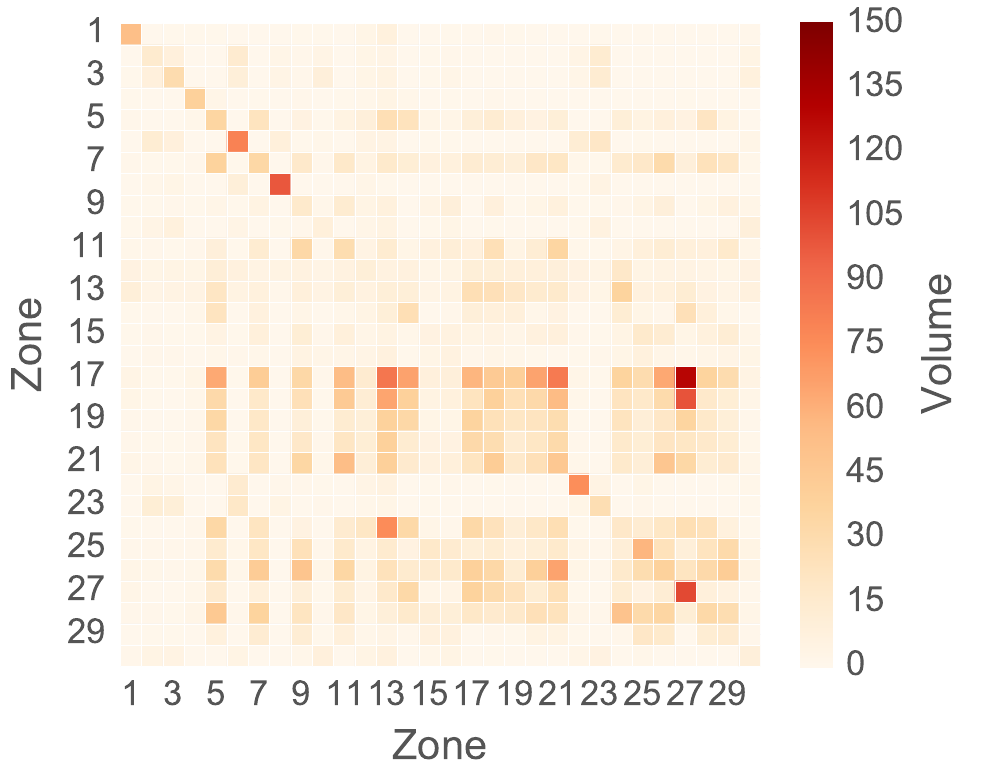}
}
\caption{Examples of passenger flow volume matrices at two time slots. We show the predicted volume using BTTF under the actual observations and 40\% NM data. Note that panels in the first row show the results during 8:00 a.m.---9:00 a.m. of June 27, and the second row corresponds to the time interval of 9:00 a.m.---10:00 a.m. of June 27.}
\label{prediction_example}
\end{figure*}

\noindent\textbf{Experiment setup}. Similar to the analyses on BTMF, we also design two missing data scenarios: random missing (RM) by randomly removing entries in the tensor and non-random missing (NM) by randomly selecting certain amount of pick-up$\times$drop-off$\times$day (or grid$\times$grid$\times$year) combinations and for each of them removing the corresponding 24h block (or 12-month block) entirely. We examine two missing rates (40\% and 60\%) and use the last seven days (i.e., $7\times 24$ time slots) and the last ten years (i.e., $10\times 12$ time slots) as the prediction periods for data (N) and (P), respectively. The time horizons of each rolling prediction are set as $\{2,4,6\}$. For tensor models, we use third-order tensor as input. The low rank of all the tensor factorization models is set as $R=30$ for both imputation and prediction tasks.

\noindent\textbf{Results and analysis}. Table~\ref{BTTF_imputation} gives the imputation performance of all models on both two data sets. Essentially, BTTF achieves competitive imputation results among these tensor-based models. For data set (N), BTTF and BTRTF perform better than other models in most cases. For data set (P), HaLRTC outperforms other models. A possible reason is that data (P) demonstrates clear low-rank structure with only a few dominating factors, which makes the nuclear norm-based method more powerful. BTTF and BTRTF show comparable performance on both data sets.

Table~\ref{bttf_forecasting_result} shows the results of multi-step prediction by using BTTF and BTRTF with certain rates of missing values. As can be seen, BTTF performs much better than BTRTF on the prediction tasks. Again, the result verifies the superiority of the VAR prior in BTTF. To demonstrate the prediction results of BTTF, we depict the actual and predicted values for six selected demand time series in Fig.~\ref{nyc_prediction_example_time_series}. As we can see, the prediction results are quite good even after a full day of missing data. To have a global overview, we plot two examples of passenger flow volumes at two time slots in Fig.~\ref{prediction_example}. The full demand matrix can be well predicted/reproduced with the tensor representation. Taken together, these results suggest that the temporal trend and overall flow rate are well characterized by the proposed BTTF model, even in severe missing conditions (e.g., predicted values in Fig.~\ref{nyc_prediction_example_time_series}(b)).

\section{Conclusion and Future Work} \label{sec:conclusion}

In this paper we present a Bayesian Temporal Factorization (BTF) framework by incorporating a VAR layer into traditional Bayesian probabilistic matrix/tensor factorization algorithms. The integration allows us to better model the complex temporal dynamics and covariance structure of multidimensional time series data on the latent dimension. By diagnosing the coefficient matrices in VAR, one can identify and interpret important causal relationships among different temporal factors. Therefore, BTF provides a powerful tool to handle incomplete/corrupted time series data for both imputation and prediction tasks. In addition, the Bayesian scheme allows us to estimate the posterior distribution of target variables, which is critical to risk-sensitive applications. For model inference, we derive an efficient and scalable Gibbs sampling algorithm by introducing conjugate priors. The fully Bayesian treatment offers additional flexibility in terms of parameter tuning while avoiding overfitting issues. We examine the framework on several real-world time series matrices/tensors, and BTF framework has demonstrated superior performance over other baseline models. Although we introduce BTF in a spatiotemporal setting, the model can be readily applied on general multidimensional time series data.

There are several directions to explore for future research. First, we can extend this framework by including a rank determination component, thus learning the number of  latent factors (i.e., rank) instead of defining it in advance \cite{zhao2015bayesianCP}. Second, we can extend this framework to account for spatial dependencies/correlations by incorporating tools such as spatial AR and Gaussian process structures. For example, when spatial information is available, we can place Gaussian process priors on the rows on $W$ instead of the independent Gaussian prior on each column \cite{luttinen2009variational,zhou2012kernelized}. Similarly, temporal factors can be also modeled as Gaussian processes as an alternative to VAR \cite{luttinen2009variational,damianou2011variational}. Third, the graphical model can be further enhanced by accommodating exogenous variables and modeling other distributions beyond Gaussian (e.g., Poisson data  \cite{gopalan2014bayesian,pmlr-v48-schein16}). To overcome the impacts of outliers, the framework can also be transformed to robust models for non-Gaussian noise \cite{luttinen2012bayesian,zhao2015bayesian}. We would like to integrate recent advances in Bayesian particle filtering and deep learning to better capture the complex and non-linear dynamics on temporal latent factors (see, e.g., \cite{arulampalam2002tutorial,rangapuram2018deep,wang2019deep}).


\section*{Acknowledgements}
This research is supported by the Natural Sciences and Engineering Research Council (NSERC) Discovery Grant, the Fonds de recherche du Quebec – Nature et technologies (FRQNT) New University Researchers Start-up Program, the Canada Foundation for Innovation (CFI) John R. Evans Leaders Fund. This project was also partially funded by IVADO Fundamental Research Project. Xinyu Chen would like to thank the Institute for Data Valorization (IVADO) for providing PhD scholarship. The numerical experiment of this study is available at \url{https://github.com/xinychen/transdim}.




\bibliographystyle{IEEEtran}


\bibliography{reference}
%



%

\begin{IEEEbiography}[{\includegraphics[width=1in,height=1.25in,clip,keepaspectratio]{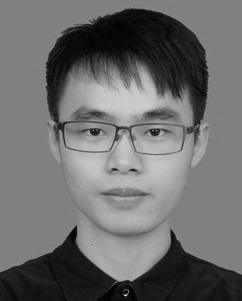}}]{Xinyu Chen} received the B.S. degree in Traffic Engineering from Guangzhou University, Guangzhou, China, in 2016, and M.S. degree in Transportation Information Engineering \& Control from Sun Yat-Sen University, Guangzhou, China, in 2019. He is currently a PhD student with the Department of Civil, Geological and  Mining Engineering, Polytechnique Montreal, Montreal, QC, Canada. His current research centers on spatiotemporal data modeling and intelligent transportation systems.
\end{IEEEbiography}

\begin{IEEEbiography}[{\includegraphics[width=1in,height=1.25in,clip,keepaspectratio]{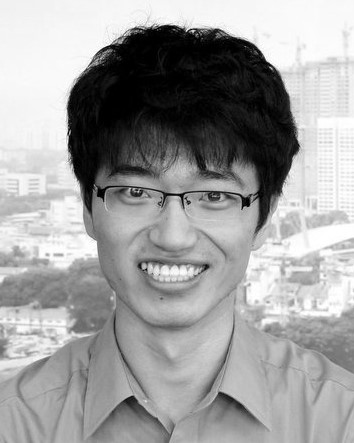}}]{Lijun Sun} received the B.S. degree in Civil Engineering from Tsinghua University, Beijing, China, in 2011, and Ph.D. degree in Civil Engineering (Transportation) from National University of Singapore in 2015. He is currently an Assistant Professor with the Department of Civil Engineering at McGill University, Montreal, QC, Canada. His research centers on intelligent transportation systems, machine learning, spatiotemporal modeling, travel behavior, and agent-based simulation. He is a member of the IEEE.
\end{IEEEbiography}







\end{document}